\RequirePackage{fix-cm}
\documentclass[twocolumn]{svjour3}          
\smartqed  
\usepackage{mathptmx}      
\usepackage[export]{adjustbox}
\usepackage{latexsym}
\usepackage{ifsym}
\usepackage[toc,acronym,nonumberlist,numberline,shortcuts]{glossaries}

\usepackage[labelfont=bf]{subcaption}   
\DeclareCaptionLabelSeparator{nspace}{\hspace{4mm}} 
\usepackage[labelfont=bf,labelsep=nspace]{caption}
\usepackage{rotating}
\usepackage{placeins}     
\usepackage{wrapfig}    
\usepackage{float}      
\usepackage[authoryear]{natbib}
\usepackage{mathtools} 
\usepackage{amssymb}
\usepackage{mdwlist}        
\usepackage{paralist}
\usepackage{url}
\usepackage[bookmarks,breaklinks=true]{hyperref}
\hypersetup{colorlinks=true,
allcolors=blue,
pdfpagemode=UseOutlines,
bookmarksdepth=3}
\usepackage{hypcap}
\usepackage[nottoc,numbib]{tocbibind}
\newcommand{\figref}[1]{Fig.~\ref{#1}}

\newcommand{\secref}[1]{Sect.~\ref{#1}} 
\newcommand{\eqnref}[1]{Eq.~\ref{#1}} 

\DeclarePairedDelimiter{\abs}{\lvert}{\rvert}

\mathcode`\*="8000
{\catcode`\*\active\gdef*{\cdot}}

\journalname{Autonomous Robots}
\newacronym{FOR}{f.o.r.}{frame of reference}
\newacronym{GT}{GT}{Ground-Truth}
\newacronym{1D}{1D}{1-Dimensional}
\newacronym{2D}{2D}{2-Dimensional}
\newacronym{3D}{3D}{3-Dimensional}
\newacronym{MSL}{MSL}{Mean Sea Level}
\newacronym{DOF}{DOF}{Degrees of Freedom}
\newacronym{CG}{CG}{Center of Gravity}
\newacronym{NED}{NED}{North-East-Down}
\newacronym{SD}{SD}{Standard Deviation}
\newacronym{UWB}{UWB}{Ultra Wide-Band}
\newacronym{MCS}{MCS}{Motion Capture System}

\newacronym{ER}{ER}{Evolutionary Robotics}
\newacronym{GA}{GA}{Genetic Algorithm}
\newacronym{ACO}{ACO}{Ant Colony Optimization}
\newacronym{PSO}{PSO}{Particle Swarm Optimization}
\newacronym{FSM}{FSM}{Finite State Machine}
\newacronym{PFSM}{PFSM}{Probabilistic Finite State Machine}
\newacronym{NN}{NN}{Neural Network}
\newacronym{RNN}{RNN}{Recurrent Neural Network}
\newacronym{FNN}{FNN}{Feed-forward Neural Network}
\newacronym{BT}{BT}{Behavior Tree}
\newacronym{RL}{RL}{Reinforcement Learning}

\newacronym{LS}{LS}{Least Squares}
\newacronym{RLS}{RLS}{Recursive Least Squares}
\newacronym{KF}{KF}{Kalman Filter}
\newacronym{EKF}{EKF}{Extended Kalman Filter}
\newacronym{UKF}{UKF}{Unscented Kalman Filter}
\newacronym{PF}{PF}{Particle Filter}
\newacronym{IAEKF}{IAEKF}{Iterative Adaptive EKF}
\newacronym{KCF}{KCF}{Kalman Consensus Filter}
\newacronym{LPF}{LPF}{Low-Pass Filter}
\newacronym{BPF}{BPF}{Band-Pass Filter}
\newacronym{HPF}{HPF}{High-Pass Filter}
\newacronym{MAF}{MAF}{Moving Average Filter}

\newacronym{CRR}{CRR}{Conflict Resolution Rate}
\newacronym{FT}{FT}{Flight Time}
\newacronym{RMSE}{RMSE}{Root Mean Squared Error}
\newacronym{ZMGN}{ZMGN}{Zero-Mean Gaussian Noise}

\newacronym{GPS}{GPS}{Global Positioning System}
\newacronym{IR}{IR}{Infra-Red}
\newacronym{IMU}{IMU}{Inertial Measurement Unit}
\newacronym{SLAM}{SLAM}{Simultaneous Localization and Mapping}

\newacronym{AOA}{AOA}{Angle of Arrival}
\newacronym{TOA}{TOA}{Time of Arrival}
\newacronym{TDOA}{TDOA}{Time Difference of Arrival}
\newacronym{RTOA}{RTOA}{Round-trip Time of Arrival}

\newacronym{WSN}{WSN}{Wireless Sensor Network}
\newacronym{WLAN}{WLAN}{Wireless Local Area Network}
\newacronym{RSS}{RSS}{Received Signal Strength}
\newacronym{RSSI}{RSSI}{Received Signal Strength Indication}
\newacronym{FSL}{FSL}{Free Space Loss}
\newacronym{BLE}{BLE}{Bluetooth Low Energy}
\newacronym{GRPR}{GRPR}{Golden Receiver Power Range}
\newacronym{ISM}{ISM}{Industrial, Scientific and Medical}
\newacronym{AP}{AP}{Access Point}
\newacronym{MAC}{MAC}{Media Access Control}
\newacronym{IoT}{IoT}{Internet of Things}
\newacronym{LD}{LD}{Log-Distance}
\newacronym{LQI}{LQI}{Link Quality Indicator}

\newacronym{SQC}{SQC}{Sum Quadratic Constraint}
\newacronym{RANSAC}{RANSAC}{RANdom SAmpling and Consensus}
\newacronym{RGB}{RGB}{Red-Green-Blue}
\newacronym{LED}{LED}{Light-Emitting Diode}
\newacronym{LoG}{LoG}{Laplacian of Gaussian}
\newacronym{SIFT}{SIFT}{Scale-Invariant Feature Transform}
\newacronym{SURF}{SURF}{Speeded Up Robust Feature}
\newacronym{OF}{OF}{Optical Flow}
\newacronym{FAST}{FAST}{Features from Accelerated Segment Test}
\newacronym{CenSurE}{CenSurE}{Center Surround Extremas for Realtime Feature Detection and Matching}

\newacronym{CC}{CC}{Collision Cone}
\newacronym{VO}{VO}{Velocity Obstacle}
\newacronym{RVO}{RVO}{Reciprocal Velocity Obstacle}
\newacronym{HRVO}{HRVO}{Hybrid Reciprocal Velocity Obstacle}
\newacronym{ORCA}{ORCA}{Optimal Reciprocal Collision Avoidance}
\newacronym{HL}{HL}{Human-Like}
\newacronym{CALU}{CALU}{Collision Avoidance under Localization Uncertainty}
\newacronym{COCALU}{COCALU}{Convex Outline Collision Avoidance under Localization Uncertainty}

\newacronym{ROS}{ROS}{Robotics Operating System}
\newacronym{SIDPAC}{SIDPAC}{System Identification Programs for Aircraft}
\newacronym{STDMA}{STDMA}{Self-Organized Time Division Multiple Access}

\newacronym{UAV}{UAV}{Unmanned Air Vehicle}
\newacronym{MAV}{MAV}{Micro Air Vehicle}

\begin{document}

\title{On-board Communication-based Relative Localization for Collision Avoidance in Micro Air Vehicle teams}

\titlerunning{On-board Communication-based Relative Localization for Collision Avoidance in Micro Air Vehicle teams}

\author{Mario Coppola$^{1,2}$ \and
        Kimberly N. McGuire$^1$ \and \\
        Kirk Y.W. Scheper$^1$ \and
        Guido C.H.E. de Croon$^1$
}

\authorrunning{M. Coppola \and
        K.N. McGuire \and 
        K.Y.W. Scheper \and 
        G.C.H.E. de Croon}

\institute{M. Coppola \at
              \email{\href{mailto:m.coppola@tudelft.nl}{m.coppola@tudelft.nl}}
           \and
           K.N. McGuire \at
				   		\email{\href{mailto:k.n.mcguire@tudelft.nl}{k.n.mcguire@tudelft.nl}}            
           \and
           K.Y.W. Scheper \at
				   		\email{\href{k.y.w.scheper@tudelft.nl}{k.y.w.scheper@tudelft.nl}}
           \and
           G.C.H.E de Croon \at	
				   		\email{\href{g.c.h.e.decroon@tudelft.nl}{g.c.h.e.decroon@tudelft.nl}}\\ \newline
        ${^1}$Delft University of Technology, Faculty of Aerospace Engineering \\
        \emph{Department of Control and Simulation (Micro Air Vehicle Laboratory)}.
        Kluyverweg 1, 2629HS, Delft, The Netherlands.\\
        ${^2}$Delft University of Technology, Faculty of Aerospace Engineering \\
        \emph{Department of Space Systems Engineering}.
        Kluyverweg 1, 2629HS, Delft, The Netherlands. \\
}

\date{Received: 14 February 2017 / Accepted: date}

\maketitle

\begin{abstract}
  Micro Air Vehicles (MAVs) will unlock their true potential once they can operate in groups.
  To this end, it is essential for them to estimate on-board the relative location of their neighbors.
  The challenge lies in limiting the mass and processing burden needed to enable this.
  We developed a relative localization method that only requires the MAVs to communicate via their wireless transceiver.
  Communication allows the exchange of on-board states (velocity, height, and orientation), while the signal-strength provides range data.
  These quantities are fused to provide a full relative location estimate.
  We used our method to tackle the problem of collision avoidance in tight areas.
  The system was tested with a team of AR.Drones flying in a 4m$\mathbf{\times}$4m area and with miniature drones of $\approx50g$ in a 2m$\mathbf{\times}$2m area.
  The MAVs were able to track their relative positions and fly several minutes without collisions.
  Our implementation used Bluetooth to communicate between the drones.
  This featured significant noise and disturbances in signal-strength, which worsened as more drones were added.
  Simulation analysis suggests that results can improve with a more suitable transceiver module.
  \keywords{Relative Localization
  \and Collision Avoidance
  \and Micro Air Vehicles
  \and Autonomous Flight
  \and Indoor Exploration
  \and Swarm}
\end{abstract}


\section{Introduction}
\label{sec:introduction}
  The agility and small scale of \glspl{MAV} make them ideal for indoor exploration \citep{kumar2012opportunities}.
  We imagine several autonomous MAVs navigating through a building/house for mapping or inspection.
  The swarm could spread out and thus complete the task in a short time.
  This approach would also bring robustness, scalability, and flexibility to the system, being no longer tied to the success and ability of only one unit \citep{brambilla2013swarm}.
  During this scenario, however, it may happen that a few MAVs end up flying together in a small area (e.g. a common bedroom, office, meeting room, hallway), leading to a significant risk of intra-swarm collisions \citep{thesistamas}.
  This is a failure condition to be avoided to ensure mission success without the unwanted loss of units.
  We have developed and tested a method to tackle this issue which uses only wireless communication between MAVs.
  Two or more MAVs estimate their relative location via the wireless connection and adjust their path to avoid collisions.
  In this paper, we describe the details of the algorithm and present real-world results on autonomous drones.\\

  \emph{The primary contribution in this article is an on-board relative localization method for MAVs based on intra-swarm wireless communication}.
  The communication channel is used as a method for the exchange of own state measurements \emph{and} as a measure of relative range (based on signal strength), this provide each MAV sufficient data to estimate the relative location of another.
  Our implementation uses Bluetooth, which is readily available at a low mass, power, and cost penalty even on smaller MAVs \citep{mcguire2016local}.
  The advantages of our solution are:
  \begin{inparaenum}[a)]
    \item it provides direct MAV-to-MAV relative location estimates at all relative bearings;
    \item it has a low dependence on the lighting and sound conditions of the environment;
    \item it has low mass, battery, and processing requirements;
    \item it does not require the use of dedicated sensors.
  \end{inparaenum}
  The findings also apply to other indoor localization problems, because it shows that only one access point is sufficient to obtain a localization estimate, as opposed to multiple ones in current state of the art \citep{malyavej2013indoor,choudry2017rssi}.
  \emph{The secondary contribution is a reactive collision avoidance strategy that is easily tailored to the anticipated performance of the localization estimates}.
  The strategy was inspired by the concept of collision cones \citep{fiorini1998motion}, tailored to suit the expected relative localization performance.\\

  The paper is organized as follows. 
  First, we review a set of related literature in \secref{sec:relatedwork}, exploring other approaches towards our goal.
  Then \secref{sec:relativelocalization} introduces the communication-based relative localization methodology, 
  and \secref{sec:avoidancebehavior} describes our collision avoidance strategy.
  To test the system, we developed a representative room exploration task, explained in \secref{sec:testsetup}.
  We gradually detail the experiments and results that have been performed, starting from simulation (\secref{sec:simulationexperiments}) to real-world fully autonomous drones (\secref{sec:realworldexperiments_autonomous}), with positive results.
  All results are further discussed in \secref{sec:discussion}.
  Concluding statements and future challenges are laid out in \secref{sec:conclusion}.

\section{Related Work and Research Context}
\label{sec:relatedwork}
  MAVs should be designed to be as efficient as possible to decrease mass and maximize flight-time.
  This means that they are often limited in sensing, computational power, and payload capabilities \citep{remes2014lisa,mulgaonkar2015design}.
  Collision avoidance is important for mission success but it must not exhaust the already limited resources, which should remain free to pursue the real mission.
  Arguably, the simplest method to avoid collisions is to have the MAVs fly at different heights.
  However, experiments by \cite{powers2013influence} have shown that MAV multi-rotors flying on top of each-other are subject to considerable aerodynamic disturbances.
  Furthermore, height sensor (e.g. sonar) readings could be disturbed.
  Based on this limitation, we conclude that lateral evasive maneuvers are needed, and these require relative location estimates between MAVs. \\

  One method to achieve relative localization is to provide a shared reference frame in which each \gls{MAV} knows its own absolute location.
  The MAVs can share absolute pose data and infer a relative estimate.
  In outdoor tasks, \gls{GPS} receivers can be used to obtain global position data to share.
  This has enabled formation flying \citep{min2016formation} and large-scale flocking \citep{vasarhelyi2014outdoor}.
  In indoor tasks, where GPS is not available, absolute position data can be measured using external sensors/beacons in a known configuration, such as: 
  motion tracking cameras \citep{michael2010grasp},
  fixed wireless transmitters/receivers \citep{guo2016ultrawideband,ledergerber2015robot},
  or visual markers \citep{faigl2013low}.
  However, these solutions are unsuitable for indoor exploration tasks because they rely on a pre-arranged environment.
  \gls{SLAM} methods circumvent this by generating an indoor map on-board during flight, which then provide position information that can be shared \citep{scaramuzza2014vision}. 
  However, if on-board map generation is not part of the mission, then this is a resource intensive practice to be discouraged \citep{ho2015optical}.
  Therefore, the more direct strategy is for the MAVs to directly localize each-other. \\

  To this end, vision has received significant attention, where front-facing cameras are used to detect and localize other MAVs.
  Current implementations generally adopt mounted visual aids in the form of:
  colored balls \citep{roelofsen2015reciprocal},
  tags \citep{conroy20143},
  or markers \citep{nageli2014environment}.
  However, experiments during exploratory phases of this study have shown that using vision without such aids, for very small drones, and at low resolution (128$\times$96px, as seen on the Lisa-S Ladybird \citep{mcguire2016local}) is prone to false-positives/false-negatives.
  Other disadvantages of using vision are: 
  dependence on lighting conditions, 
  the need for a front-facing camera, 
  limited field-of-view, 
  and high processing requirements \citep{alvarez2016collision}. \\

  \cite{roberts20123} proposed \gls{IR} sensors.
  If arranged in an array, these enables an accurate measure or relative bearing between two MAVs.
  Unfortunately, because \gls{IR} is uni-directional, several sensors are needed to each face in a direction.
  This is not easily exportable to smaller MAVs. \\

  Alternatively, recent work by \cite{basiri2015audioPHDTHESIS} uses on-board sound-based localization.
  A microphone array and a chirp generator are mounted on-board of the MAVs, and the difference between arrival times of the chirp at the different microphones is used to estimate the relative bearing \citep{basiri2014audio, basiri2016board}.
  This method requires dedicated hardware, which for smaller MAVs can account for an increase in mass of even 10\%-20\% \citep{basiri2016board, remes2014lisa}. \\

  To truly minimize the footprint, we decide to focus on a component that is mounted by necessity on all MAVs: 
  a wireless transceiver.
  This is typically used for communication with a ground station \citep{lehnert2013muav, mcguire2016local},
  but it may also be used for intra-swarm communication.
  The signal strength of a wireless communication decreases with distance from the antenna, and can be used as a measure for range between MAVs.
  \cite{thesistamas} first exploited this on-board of real MAVs as a measure for range sensing.
  However, range-only data, coupled with significant noise and disturbances, were found insufficient to guarantee safe flight of two or more MAVs in a confined area in-spite of relying on a complex evolved avoidance behavior.
  \cite{amrita2016bluetooth} recently also explored range-only avoidance on WeBot robots (in simulation only), but range measurements were aided by an array of proximity sensors. \\

  Transceivers can be exploited for both ranging and data-exchange.
  Based on this, we developed a fusion filter that can determine relative location estimates via communicating on-board states between MAVs.
  To the best of our knowledge, the only instance of on-board relative localization using a wireless transceiver was recently brought forward by \cite{guo2016relative} with \gls{UWB} technology.
  However, they make use of one of the MAVs as a static beacon and their method relies on highly accurate distance measurements.
  Instead, we propose a method that complements possibly noisy distance measurements by communicating on-board states between \emph{moving} MAVs.
  We then show how it can be used for indoor collision avoidance.
  We extensively validate this on real platforms as light as 50g that communicate between each other using Bluetooth, which is highly prone to noise and disturbances.

\section{Communication-Based Relative Localization}
\label{sec:relativelocalization}  
  Relative localization is achieved via wireless communication link between the MAVs.
  The idea is that the MAVs communicate the following states to each-other:
  planar velocity in the body frame,
  orientation with respect to North,
  and height from the ground.
  When communicating, the MAVs can also capture the strength of the signal, which acts as a measure of distance.
  For \gls{BLE}, the technology chosen in our implementation, signal-strength measurements are referred to as \gls{RSSI}.
  Each MAV fuses the received states, the RSSI, and its own on-board states to estimate the relative pose of another MAV.
  When multiple MAVs are present, multiple instances of the fusion filter run in parallel so that each MAV may keep track of the others.  
  This section details the design and implementation of the relative localization scheme and presents some preliminary localization results that were obtained in early stages of the research.

\subsection{Framework Definition for Relative Localization}
\label{sec:framework}
\begin{figure}[t]
    \centering
    \includegraphics[width=84mm]{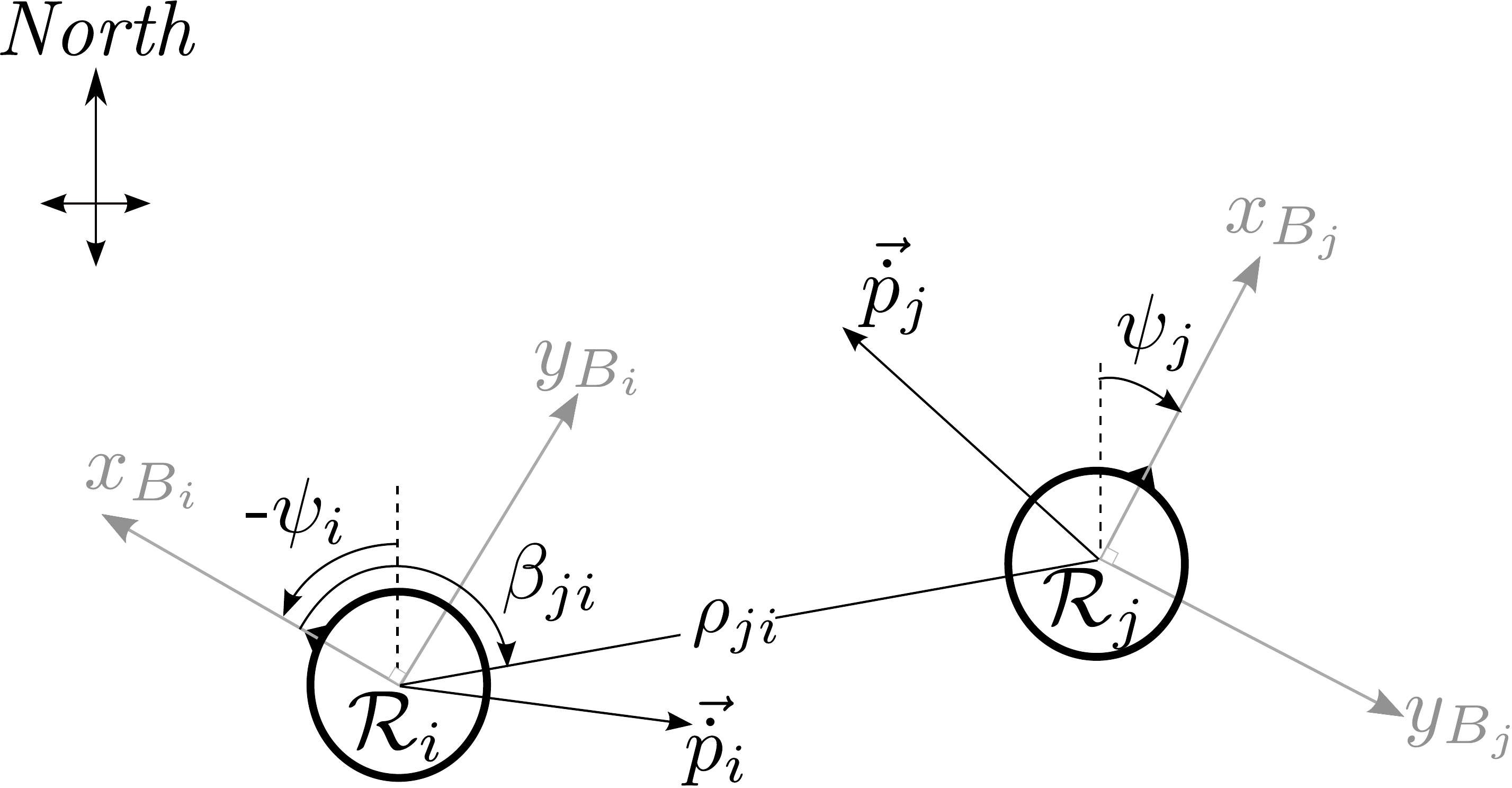}
    \caption{Top view of the relative localization framework ($x_B$ and $y_B$ are the planar axis of $\mathcal{F}_{B}$, while $z_B$ is positive down-wards)
    }
    \label{fig:framework}
  \end{figure}

  Consider two MAVs $\mathcal{R}_i$ and $\mathcal{R}_j$ with body-fixed frames $\mathcal{F}_{{B}_i}$ and $\mathcal{F}_{{B}_j}$, respectively.
  We define the relative pose of $\mathcal{R}_j$ with respect to $\mathcal{R}_i$ as the set
  $\vec{{P}}_{ji} = \begin{bmatrix} \rho_{ji}, & \beta_{ji}, & z_{ji}, & \psi_{ji} \end{bmatrix}$
  , where
  $\rho_{ji}$ represents the range between the origins of $\mathcal{F}_{{B}_i}$ and $\mathcal{F}_{{B}_j}$,
  $\beta_{ji}$ is the horizontal planar bearing of the origin of $\mathcal{F}_{{B}_j}$ with respect to $\mathcal{F}_{{B}_i}$,
  $z_{ji}$ is the height of $\mathcal{R}_j$ with respect to $\mathcal{R}_i$ and
  $\psi_{ji}$ is the yaw of $\mathcal{F}_j$ with respect to $\mathcal{F}_i$.
  See \figref{fig:framework} for an illustration.
  Note that $\rho_{ji}$ and $\beta_{ji}$ are related to their cartesian counterparts via:
  \begin{align}
    \rho_{ji} &= \sqrt{x_{ji}^2 + y_{ji}^2 + z_{ji}^2} \label{eq:rij}\\
    \beta_{ji} &= atan2(y_{ji}, x_{ji}) \label{eq:bij}
  \end{align}
  $x_{ji}$, $y_{ji}$, and $z_{ji}$ are the Cartesian coordinates of the origin of $\mathcal{R}_j$ in $\mathcal{F}_{{B}_i}$.

\subsection{Signal Strength as a Range Measurement}
\label{sec:bssasrange}

  \begin{figure*}[t]
  \centering
    \begin{subfigure}[t]{0.30\textwidth}
      \centering
      \includegraphics[width=\textwidth]{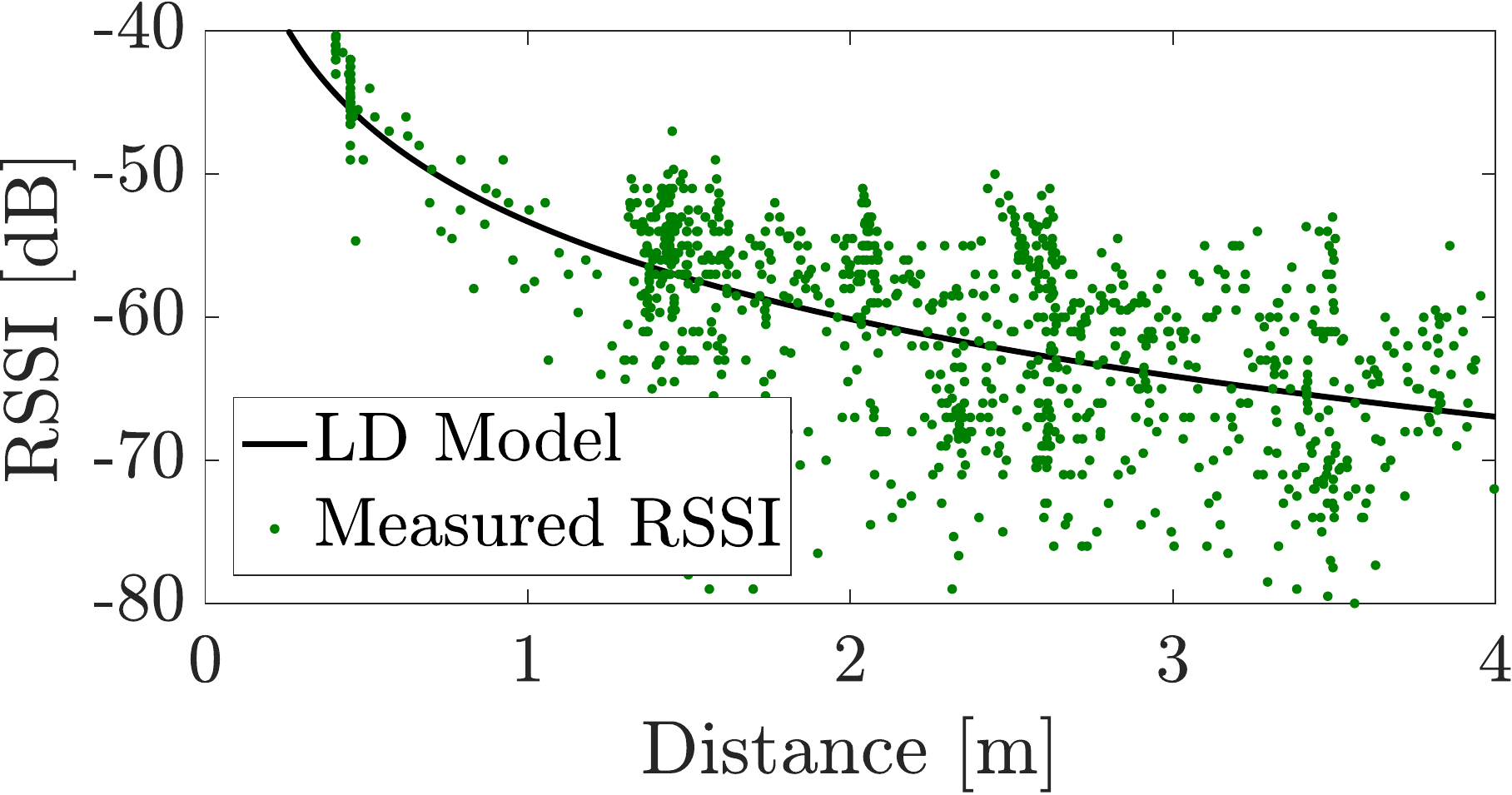}
      \caption{RSSI measurements with respect to distance (green-dotted) and fitted LD model (black, solid)}
      \label{fig:rssimodel_logdist}
    \end{subfigure}
    ~
    \begin{subfigure}[t]{0.30\textwidth}
      \centering
      \includegraphics[width=\textwidth]{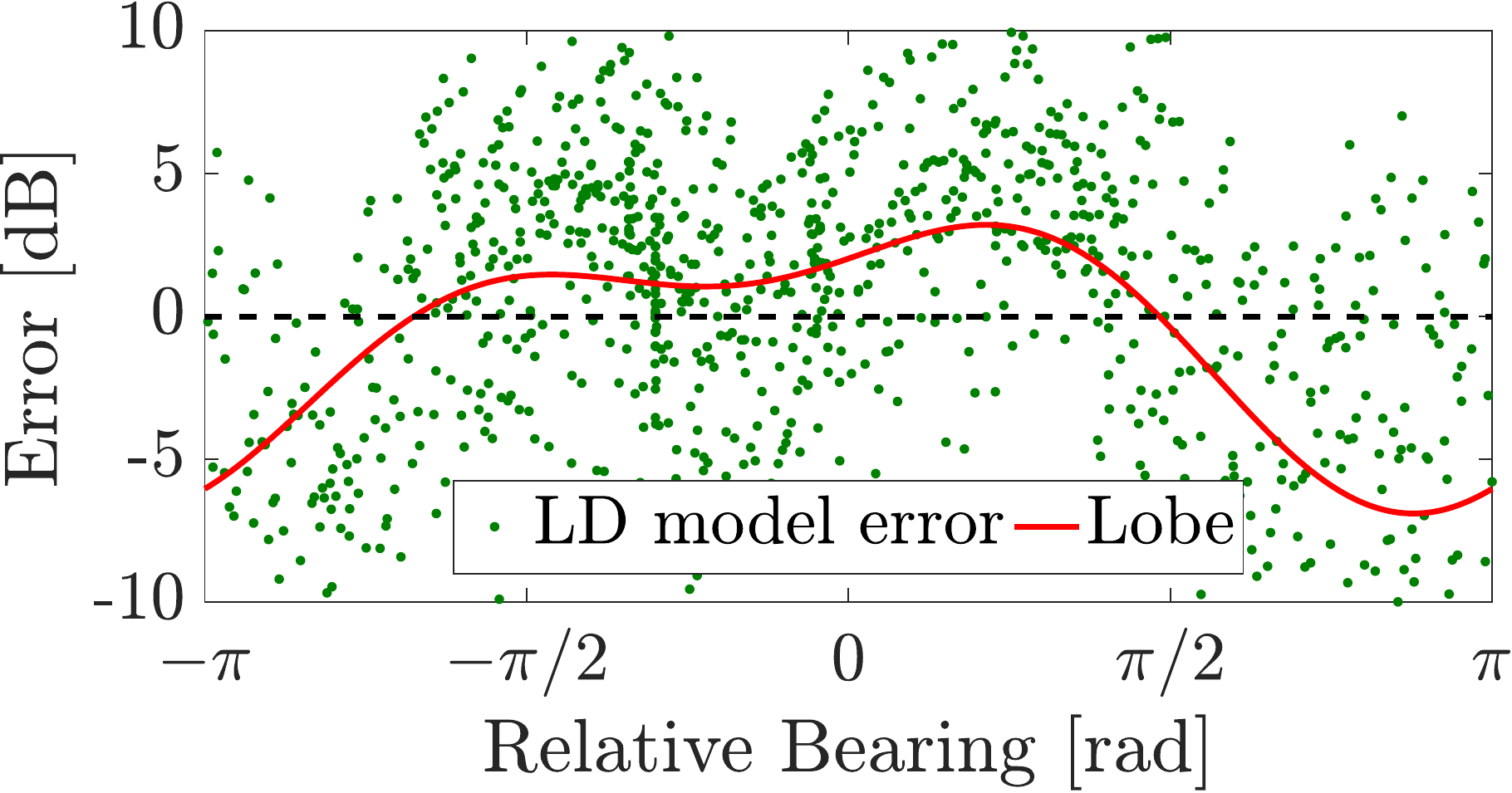}
      \caption{Error about LD model with respect to relative bearing (green, dotted) fitted with a second order Fourier series (red, solid)}
      \label{fig:rssimodel_lobes}
    \end{subfigure}
    ~
    \begin{subfigure}[t]{0.30\textwidth}
      \centering
      \includegraphics[width=\textwidth]{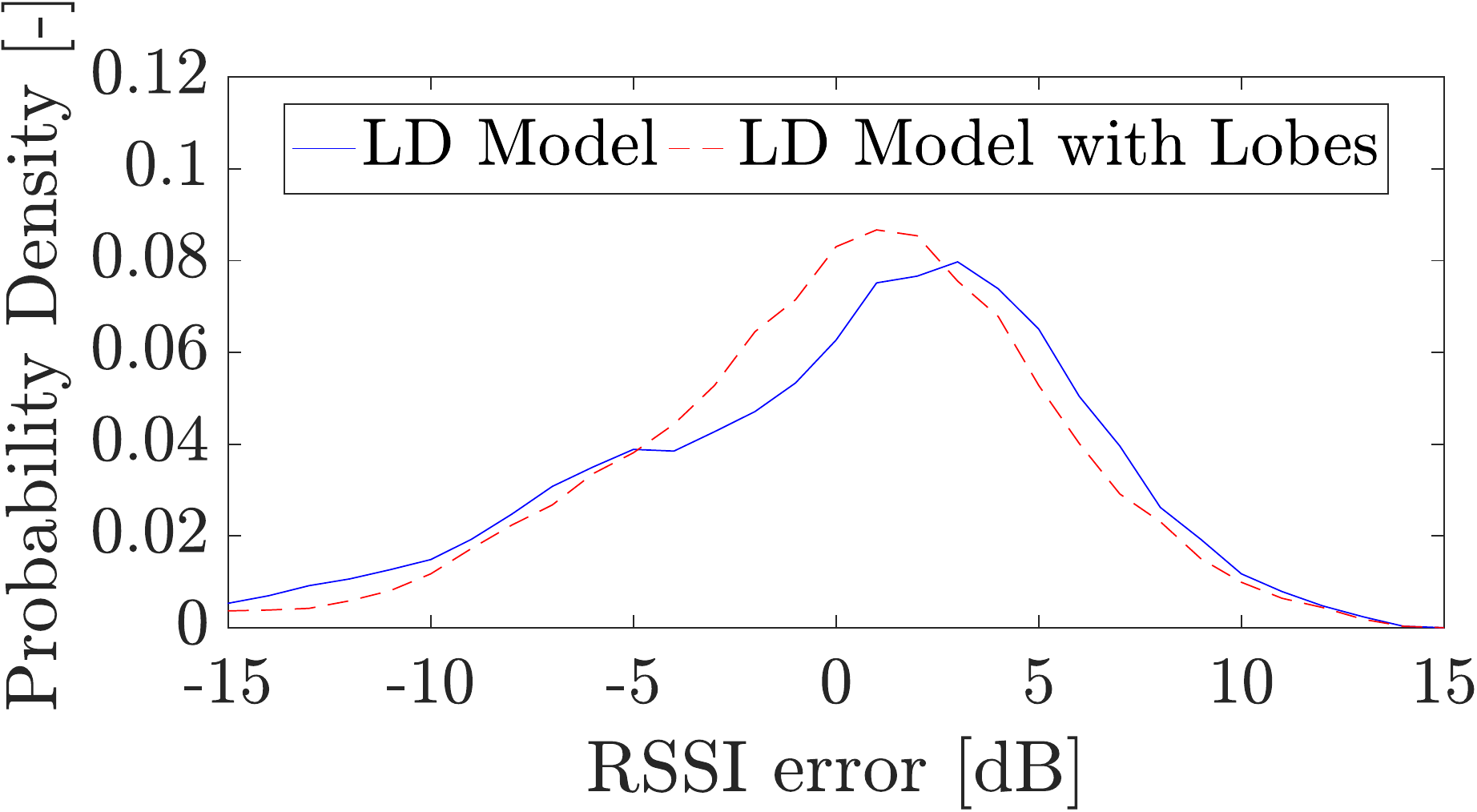}
      \caption{Noise distribution about the LD model without (blue, solid) and with (red, dashed) lobe effects}
      \label{fig:rssimodel_noise}
    \end{subfigure}
    \caption{Results of RSSI measurements during an experiment whereby a Ladybird MAV was carried in circles around a fixed Bluetooth antenna}
    \label{fig:rssimodel}
  \end{figure*}

  Let $S_{ji}$ be the \gls{RSSI} measurement in $dB$.
  It is correlated with $\rho_{ji}$ by a function $\mathcal{L}(\rho_{ji})$. 
  We define this function based on the \gls{LD} model \citep{seybold2005introduction}:
  \begin{equation}
    S_{ji} = \mathcal{L}(\rho_{ji}) = P_n - 10 * \gamma_{l} * \log_{10}\left(\rho_{ji}\right).
    \label{eq:logdistance}
  \end{equation}
  $P_n$ is the \gls{RSSI} at a nominal distance of $1m$. 
  $\gamma_{l}$ is the \emph{space-loss parameter}, which dictates how much the signal strength decays with distance (for free-space: $\gamma_{l} = 2.0$).
  \footnote{
  Experimentally, it has been found that office buildings can feature $2 \leq \gamma_l \leq 6$ \citep{kushki2008indoor}.
  Performing a sensitivity analysis of the LD model shows that an accurate identification of $\gamma_{l}$ has a low impact on the distance estimate at small distances.}
  The \gls{LD} model is assumed subject to Gaussian noise \citep{svevcko2015distance}. \\
  
  In preliminary tests, we analyzed the \gls{LD} model with Ladybird MAV \citep{remes2014lisa} connected via Bluetooth to a fixed W1049B omni-directional antenna \citep{btantenna}.
  The MAV was carried in concentric circles at different distances around the antenna whilst RSSI was being recorded with the antenna.
  Its orientation with respect to North was kept constant, thus varying the relative bearing to the antenna.
  \gls{GT} data was recorded with an Optritrack \gls{MCS}.
  The results from a representative data-sample are shown in \figref{fig:rssimodel}, to which the \gls{LD} model was fitted using a non-linear \gls{LS} estimator as in \figref{fig:rssimodel_logdist}.
  Among a set of similar experiments, the \gls{SD} of the error about the fitted LD model was found to be between $3dB$ and $6dB$.
  This is in line with literature \citep{thesistamas,nguyen2013evaluation}.\\

  We also observed a change of the error with the relative bearing.
  This is shown in \figref{fig:rssimodel_lobes}, and accounts for the skew in error distributions, see \figref{fig:rssimodel_noise}.
  The disturbances that can cause this were found to be:
  propagation lobes,
  interference by the reflection of the signal in the environment,
  the presence of other signals in the $2.4GHz$ spectrum, or
  other objects that obstruct the signal
  \citep{seybold2005introduction,svevcko2015distance,thesistamas,kushki2008indoor,caron2008indoor}.
  The LD model could be expanded with this, but the susceptibility to environmental disturbances would lead to a confounding element between bearing and range which could be detrimental to the convergence of the fusion filter.

\subsection{Localization via Fusion of Range and On-board States}
\label{sec:localizationschemeexplained}
  Achieving a relative pose estimate requires measuring or inferring all four variables in $\vec{P}_{{ji}}$.
  From those, we can directly measure or observe the following three:
  \begin{itemize*}
   \item $\rho_{ji}$:
   (range), available via \gls{RSSI} as in \secref{sec:bssasrange}.
   \item $z_{ji}$ 
   (relative height):
   Each MAV is expected to measure its height above the ground.
   This could be done with a pressure sensor \citep{beard2007state, sabatini2013stochastic, shilov2014next}, sonar, or a downward-facing camera \citep{kendoul2009adaptive, kendoul2009optic}.
   Two MAVs $\mathcal{R}_i$ and $\mathcal{R}_j$ can share their altitude data, such that: $z_{ji} = z_j - z_i$.
   \item $\Psi_{ji}$ 
    (relative orientation):
    It is assumed that all MAVs acknowledge a common planar axis, e.g. magnetic North \citep{no2015attitude, afzal2011magnetic}.
    Via communication, the MAVs can share their orientation data.
  \end{itemize*}

  Relative bearing is the only unknown variable.
  It becomes observable when fusing the three measurements above with velocity measurements, 
  as shown in \cite{martinelli2005observability} and \cite{martinelli2005multi}.
  We chose to perform sensor fusion with a discrete-time \gls{EKF} due to its efficient processing and memory requirements \citep{de2014relative}.
  The state transition model from time step $k$ to $k+1$ was defined as in \eqnref{eq:localization_process}:
  
  \begin{align}
    \begin{bmatrix}
      \vec{p}_{ji} \\
      \dot{\vec{p}}_{i} \\
      \dot{\vec{p}}_{jRi} \\
      \psi_{j} \\ 
      \psi_{i} \\
      z_{j} \\
      z_{i} \end{bmatrix}_{k+1} &= 
    \begin{bmatrix}
      \vec{p}_{ji} + \left(\dot{\vec{p}}_{jRi} - {\dot{\vec{p}}_{i}}\right) \Delta t \\
      \dot{\vec{p}}_{i} \\
      \dot{\vec{p}}_{jRi} \\
      \psi_{j} \\ 
      \psi_{i} \\
      z_{j} \\
      z_{i} 
    \end{bmatrix}_k + \vec{v}_k
    \label{eq:localization_process}
  \end{align}

  $\vec{p}_{ji} = \begin{bmatrix} x_{ji} & y_{ji} \end{bmatrix}^T$ 
  holds Cartesian equivalents of bearing and range.
  $\dot{\vec{p}}_i = \begin{bmatrix} \dot{x}_{i} & \dot{y}_{i} \end{bmatrix}^T$ 
  is a vector of the velocity of $\mathcal{R}_i$ in $\mathcal{F}_{{B}_i}$ (see \figref{fig:framework}).
  $\dot{\vec{p}}_{jRi}$ is $\dot{\vec{p}}_j$ \emph{rotated} from $\mathcal{F}_{{B}_j}$ to $\mathcal{F}_{{B}_i}$.
  $\Delta t$ is a discrete time step between updates equal to the time between $k$ and $k+1$.
  $\vec{v}_k$ represents the noise in the process at time step $k$.
  This model assumes that all current velocities and orientations remain constant between time-steps.
  The observation model for the \gls{EKF} is given by \eqnref{eq:localization_measurement}.
  \begin{align}
    \begin{bmatrix}
      S_{ji} \\
      \dot{\vec{p}}_{i} \\
      \dot{\vec{p}}_{j} \\
      \psi_{j} \\
      \psi_{i} \\
      z_j \\
      z_i \end{bmatrix}_{k} &= 
    \begin{bmatrix}
      \mathcal{L}(\rho_{ji}) \\
      \dot{\vec{p}}_{i} \\
      \mathbf{R_{2D}(\psi_{ji})} * \dot{\vec{p}}_{jRi} \\
      \psi_{j} \\
      \psi_{i} \\
      z_j \\
      z_i 
    \end{bmatrix}_k + \vec{w}_k
    \label{eq:localization_measurement}
  \end{align}
  $\mathbf{R_{2D}({*})}$ is a 2D rotation matrix that uses the relative heading $\psi_{ji}$ to rotate the state estimate $\dot{\vec{p}}_{jRi}$ from $\mathcal{F}_{B_i}$ to $\mathcal{F}_{B_j}$.
  $\vec{w}_k$ represents the noise in the measurements at time step $k$.
  Note that $\rho_{ji}$ is expanded as per \eqnref{eq:rij} so as to observe $x_{ji}$ and $y_{ji}$.\\
  The EKF cannot be initialized with a correct relative localization estimate, since this is not known; it must converge towards the correct estimate during flight.
  Appropriate tuning of the EKF noise covariance matrices is key to achieving this.
  In the EKF, the measurement noise matrix $\mathbf{R}$ is a diagonal matrix with the form shown in \eqnref{eq:noise_measurement}.
  \begin{equation}
    \mathbf{R} =
    \begin{bmatrix}
      \sigma_{m}^2 &   &  &   \\ 
        & \sigma_{v}^2 * \mathbf{I_{4\times4}} & &\\
        &  & \sigma_{\psi}^2 * \mathbf{I_{2\times2}} & \\
        &  &  & \sigma_{z}^2 * \mathbf{I_{2\times2}}\\
    \end{bmatrix}.
    \label{eq:noise_measurement}
  \end{equation}
  $\sigma_{m}$ is the assumed \gls{SD} of $S_{ji}$.
  $\sigma_{v}$ is the assumed \gls{SD} of $\dot{\vec{p}}_i$ and $\dot{\vec{p}}_j$.
  $\sigma_{\psi}$ is the assumed \gls{SD} of the magnetic orientation measurements.
  $\sigma_{z}$ is the assumed \gls{SD} of the height measurements.
  $\mathbf{I_{n\times n}}$ is a $n\times n$ identity matrix.
  Based on our preliminary \gls{RSSI} noise analysis, $\sigma_{m}$ is tuned to $5dB$.
  All other \glspl{SD} were tuned to $0.2$, unless otherwise stated.\\
  
  The process noise matrix $\mathbf{Q}$ is the diagonal matrix presented in \eqnref{eq:noise_process}.
  \begin{equation}
    \mathbf{Q} =
    \begin{bmatrix}
      \sigma_{{Q}_{p}}^2 * \mathbf{I_{2\times2}} &  &  &   \\ 
        & \sigma_{{Q}_{v}}^2 * \mathbf{I_{4\times4}} &  & \\
        &   & \sigma_{{Q}_{\psi}}^2 * \mathbf{I_{2\times2}} &  \\
        &   &    & \sigma_{{Q}_{z}}^2 * \mathbf{I_{2\times2}} \\
    \end{bmatrix}.
    \label{eq:noise_process}
  \end{equation}
  $\sigma_{{Q}_{p}}$ is the \gls{SD} of the process noise on the relative position update.
  $\sigma_{{Q}_{v}}$, $\sigma_{{Q}_{\psi}}$, and $\sigma_{{Q}_{z}}$ are \glspl{SD} for the expected updates in velocity, orientation, and height respectively.
  The tuning of $\mathbf{Q}$ defines the validity of the process equations \citep{malyavej2013indoor}.
  The tuning is made such that a \emph{high-level of trust is put on the relative position update}.
  This approach is key to encouraging convergence and helps to discard the high noise and disturbance in the RSSI measurements.
  Unless otherwise stated:
  $\sigma_{{Q}_{p}} = 0.1$, 
  while $\sigma_{{Q}_{v}} = \sigma_{{Q}_{\psi}} = \sigma_{{Q}_{z}} = 0.5$. \\
  
  The filter is limited by \emph{flip and rotation ambiguity} as defined in \citep{cornejo2015distributed}.
  When the motion of $\mathcal{R}_j$ perfectly matches the motion by $\mathcal{R}_i$, range-only measurements remain constant and are not informative for bearing estimation.
  If the MAVs do not fly in formation, the probability of this event is low \citep{cornejo2015distributed}.
  The same ambiguity takes place when both $\mathcal{R}_i$ and $\mathcal{R}_j$ are static; 
  motion by at least one MAV is required. 
  \begin{figure*}[ht]
  \centering
    \begin{subfigure}[t]{84mm}
      \centering
      \includegraphics[width=\textwidth]{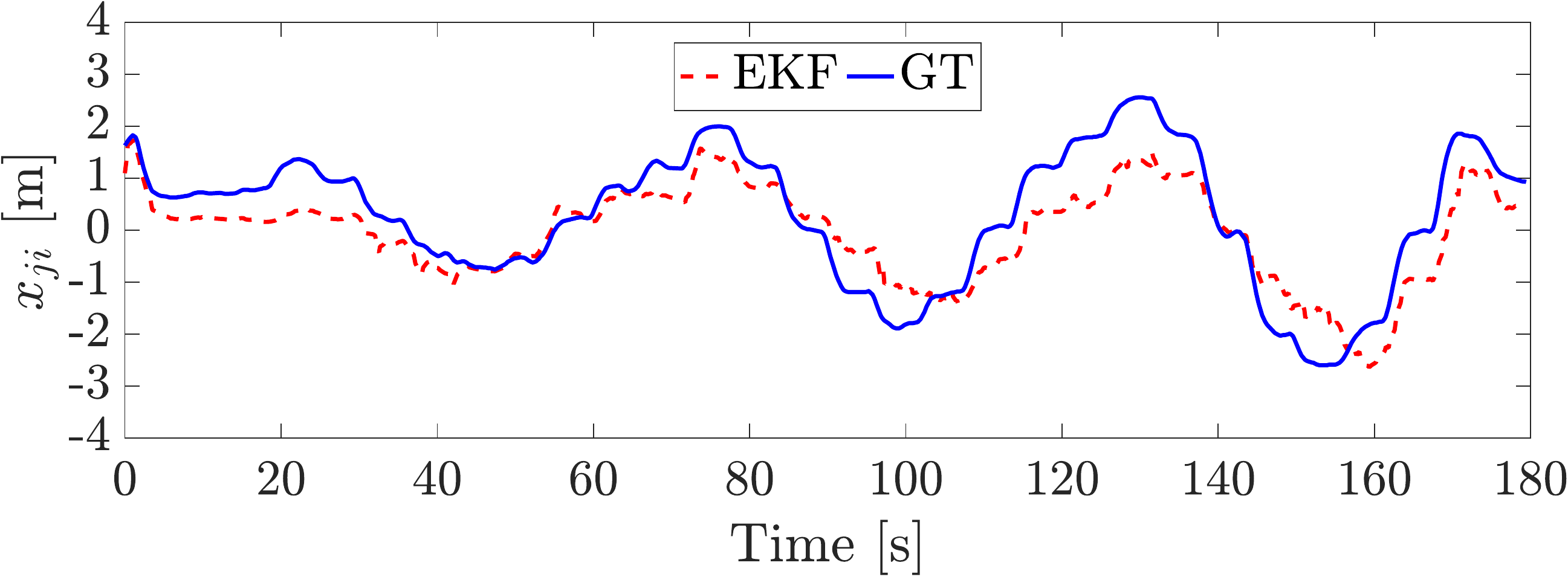}
      \caption{Ground-truth vs. estimated location of the MAV along the $x_B$ axis of the antenna}
      \label{fig:prelim_loc_x}
    \end{subfigure}
    ~
    \begin{subfigure}[t]{84mm}
      \centering
      \includegraphics[width=\textwidth]{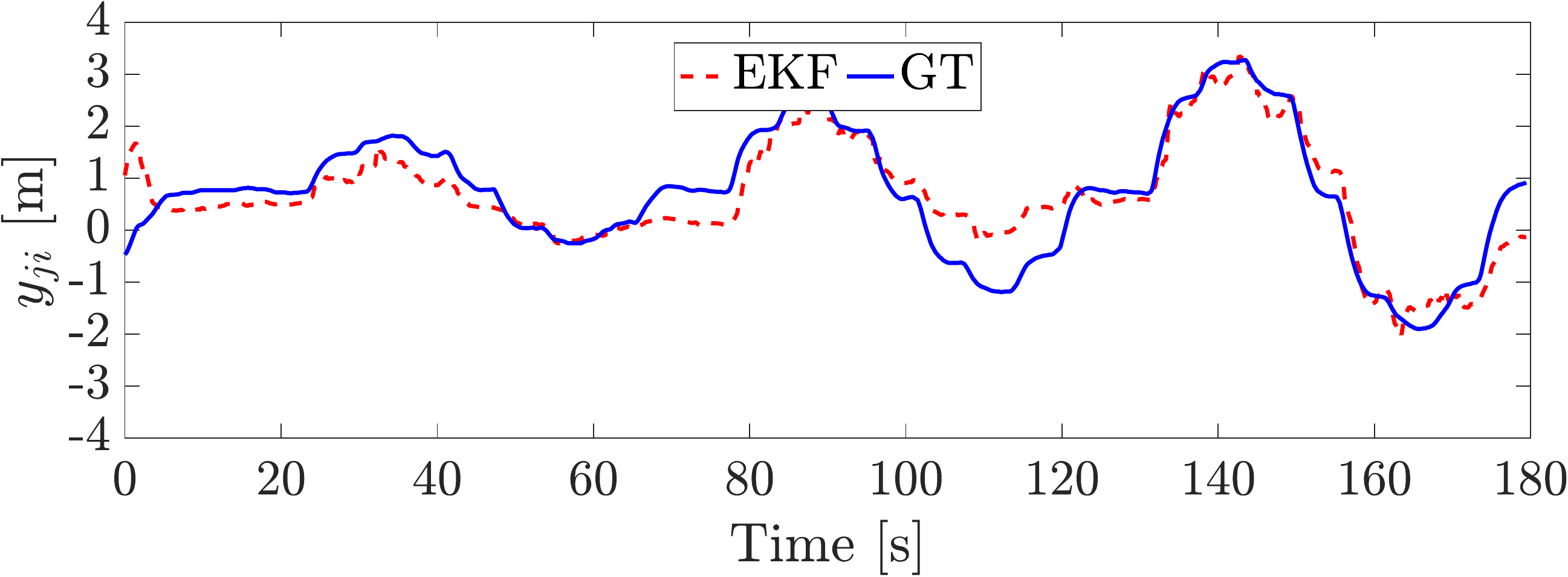}
      \caption{Ground-truth vs. estimated location of the MAV along the $y_B$ axis of the antenna}
      \label{fig:prelim_loc_y}
    \end{subfigure}
    \begin{subfigure}[t]{84mm}
      \centering
      \includegraphics[width=\textwidth]{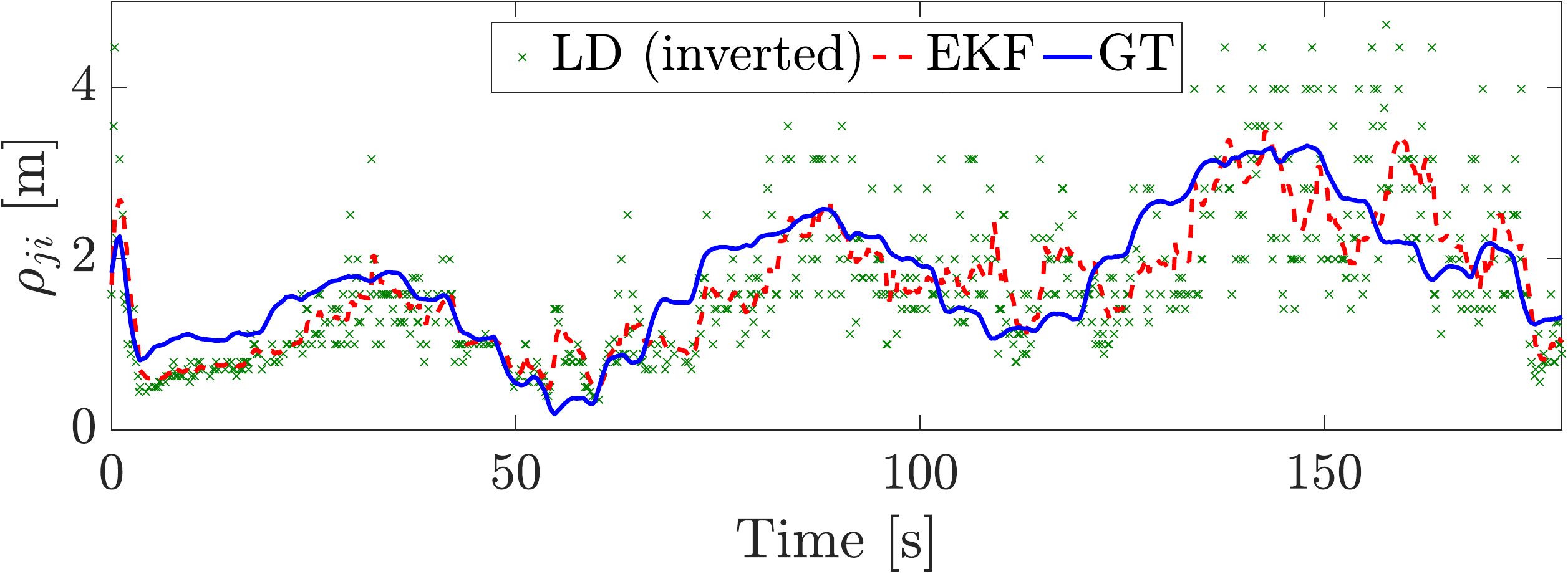}
      \caption{Comparison of EKF estimated range compared to ground truth and estimate from inverting the LD model}
      \label{fig:prelim_loc_range}
    \end{subfigure}
    ~
    \begin{subfigure}[t]{84mm}
      \centering
      \includegraphics[width=\textwidth]{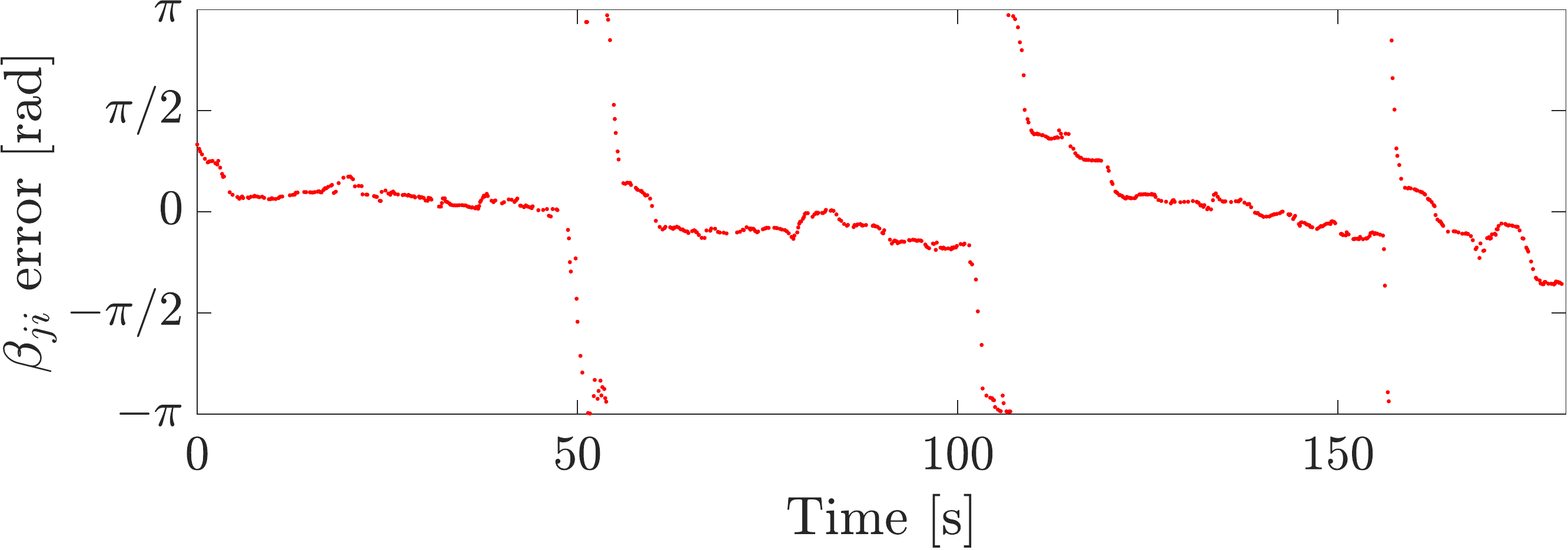}
      \caption{Error in $\beta_{ji}$ over time}
      \label{fig:prelim_loc_bearing}
    \end{subfigure}
    \caption{Preliminary localization results based on circular flights of a Ladybird MAV around a fixed antenna (averaged over 50 iterations of artificial noise added to the velocity, height, and orientation measurements)}
    \label{fig:circleflight_localization}
  \end{figure*}

\subsection{Implementation Details}
\label{sec:implementationdetailsforbluetooth}
  We used \gls{BLE} to enable communication between the MAVs.
  The data is sent and received by means of advertising messages scheduled using a \gls{STDMA} algorithm \citep{gaugel2013depth}.
  This way, each MAV's Bluetooth antenna alternates between advertising and listening.
  This enables direct communication and circumvents the Master-Slave paradigm otherwise enforced by the Bluetooth standard \citep{townsend2014getting}.
  The message rate is $5Hz$, i.e. data is received every $0.2s$.

\subsection{Preliminary Relative Localization Tests}
\label{sec:localizationperformance_preliminary}
  We performed preliminary localization tests with a Ladybird MAV around a fixed Bluetooth W1049B antenna.
  The objective was to determine how well the antenna could localize the MAV.
  An Optitrack \gls{MCS} was used to guide the MAV in circular flights and record its \gls{GT} velocity, orientation, and height.
  The antenna measured the RSSI with the MAV.
  The recorded \gls{GT} data was altered with Gaussian noise with $\sigma_{v} = 0.2m/s$, $\sigma_{z} = 0.2m$, and $\sigma_{\psi} = 0.2rad$, and then used as measurements for the EKF.
  In the LD model of the EKF: $P_n = -63dB$ and $\gamma_l = 2.0$.
  The EKF was initialized with a null guess position of $x_{ji}=y{ji}=1m$.
  In these preliminary tests, the localization filter was applied off-board.\\

  \figref{fig:circleflight_localization} shows the results.
  Estimates for $x_{ji}$ and $y_{ji}$ are shown in \figref{fig:prelim_loc_x} and \figref{fig:prelim_loc_y}, 
  the EKF converges towards GT in the first few seconds, after which it tracks successfully.
  \figref{fig:prelim_loc_range} shows the estimated range, where we can observe a significant improvement in error with respect to an inverted LD model.
  Note that the range error increases with distance, this is due to the logarithmic nature of RSSI propagation.
  \figref{fig:prelim_loc_bearing} shows the bearing error, which is small throughout most of the flight.
  The only exceptions are occasional spikes which occur at small distances which cross over $x_{ji}=0m$ and $y_{ji}=0m$.
  This is because at very small distances, a small error in $x_{ji}$ or $y_{ji}$ can translate into a significant error in $\beta_{ji}$.
  Thanks to the avoidance algorithm, such small distances should be avoided altogether during flights.

\section{Collision Avoidance Behavior}
\label{sec:avoidancebehavior}
  The avoidance algorithm was inspired by the \gls{CC} frame-work as seen in the works by \cite{fiorini1998motion} and \cite{wilkie2009generalized}.
  A collision cone is a set of all velocities of an agent that are expected to lead to a collision with an obstacle at a given point in time.
  Its name is derived from the fact that it is geometrically cone-shaped.
  This section details our implementation of collision cones and how it is used to determine an avoidance trajectory.
   
\subsection{Collision Cones and Avoidance Strategy}
\label{sec:generalcollisionavoidancestrategy}
\begin{figure}[t]
    \centering
    \includegraphics[width=84mm]{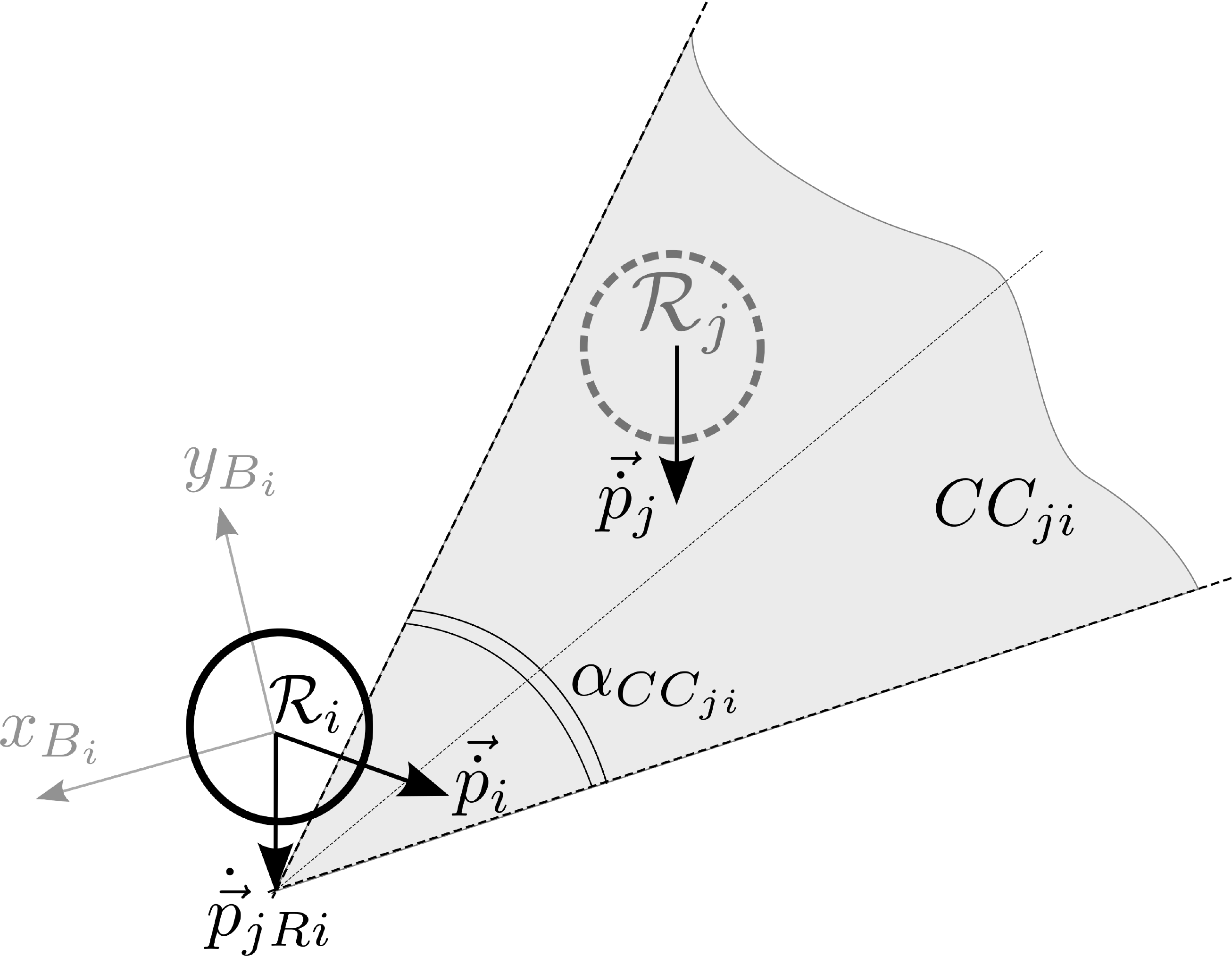}
    \caption{
    Depiction of $CC_{ji}$ that $\mathcal{R}_i$ holds with respect to the estimated location of $\mathcal{R}_j$
    }
    \label{fig:collisioncone}
  \end{figure}

  Take two MAVs $\mathcal{R}_i$ and $\mathcal{R}_j$.
  The collision cone $CC_{ji}$ (depicted in \figref{fig:collisioncone}) would include all velocities of $\mathcal{R}_i$ which could lead to a collision with $\mathcal{R}_j$.
  It is constructed in three steps.
  \begin{enumerate*}
    \item  A cone $CC_{ji}$ is defined as in \eqnref{eq:cc1}.
     $\alpha$ is an arbitrary angle.
     $x$ and $y$ are points on $x_{B_i}$ and $y_{B_i}$, respectively.
     The cone is characterized by an expansion angle $\alpha_{{CC}_{ji}}$, subject to $0 < \alpha_{{CC}_{ji}} < \pi$.
     \begin{align}
     CC_{ji} = 
       \left\{(x,y) \in \mathbb{R}^2;
       {\alpha} \in \mathbb{R};
       \abs{\alpha} \leq \frac{\abs{\alpha_{{CC}_{ji}}}}{2}
       :
       \tan(\alpha)x = y\right\} \label{eq:cc1}
     \end{align}

  \item $CC_{ji}$ is rotated so as to be centered around the estimated bearing to the obstacle $\mathcal{R}_j$ as in \eqnref{eq:cc2}, where: $\bar{\beta}_{ji}$ is the estimated ${\beta}_{ji}$ from the EKF, $\leftarrow$ is an update operator, and $\mathbf{R}(*)$ is a rotation operator for the set.
  \begin{equation}
    {CC}_{ji} \leftarrow 
    \left( \mathbf{R}(\bar{{\beta}}_{ji}) * {CC}_{ji} \right)
    \label{eq:cc2}
  \end{equation}

  \item The cone is translated by the estimated velocity of $\mathcal{R}_j$ expressed in $\mathcal{F}_{B_i}$, to account for the fact that the obstacle is moving, as per \eqnref{eq:cc3}.
  $\overline{\dot{\vec{p}}_{jRi}}$ is the estimated $\dot{\vec{p}}_{jRi}$ from the EKF.
  The operator $\oplus$ denotes the translation of a set by a vector.
  \begin{equation}
    {CC}_{ji} \leftarrow {CC}_{ji} \oplus \overline{\dot{\vec{p}}_{jRi}}
    \label{eq:cc3}
  \end{equation} 
 \end{enumerate*}

  In a team of $m$ MAVs, each member $\mathcal{R}_i$ holds $m-1$ collision cones that it can superimpose into a single set ${CC}_i$:
  \begin{equation}
    {CC}_i =\bigcup\limits_{j=1}^{m-1} {CC}_{ji}
    \label{eq:ccsuperposition}
  \end{equation}

  If, during flight, $\vec{\dot{p}}_i \in {CC}_i$, then a \emph{\textbf{clock-wise search}} about the $z_{B_i}$ axis 
  (starting with the current desired velocity) 
  is used to determine the desired escape velocity.
  If no solution is found, then the search is repeated for a higher escape speed. \\

  The clock-wise search encourages a preference for right-sided maneuvers with respect to the current flight direction.
  This differentiates it from the \gls{VO} avoidance method, which selects a flight direction that minimizes the required change in velocity.
  This automatically resolves an issue known as \emph{``reciprocal dances''}, which are left-right dances when two entities heading towards each-other repeatedly select the same escape direction.
  Other solutions to reciprocal dances assume reciprocity, meaning the assumption that the other member will also take a certain evasive action \citep{snape2009independent, snape2011hybrid, van2011reciprocal}.
  In our case, however, due to the potential for large relative localization errors, MAVs cannot safely assume that the others will participate in a suitable and reciprocal escape maneuver.

\subsection{Tuning the Expansion Angle of the Collision Cone}
\label{sec:tuningthecollisionconesize}

\begin{figure}[t]
    \centering
    \includegraphics[width=84mm]{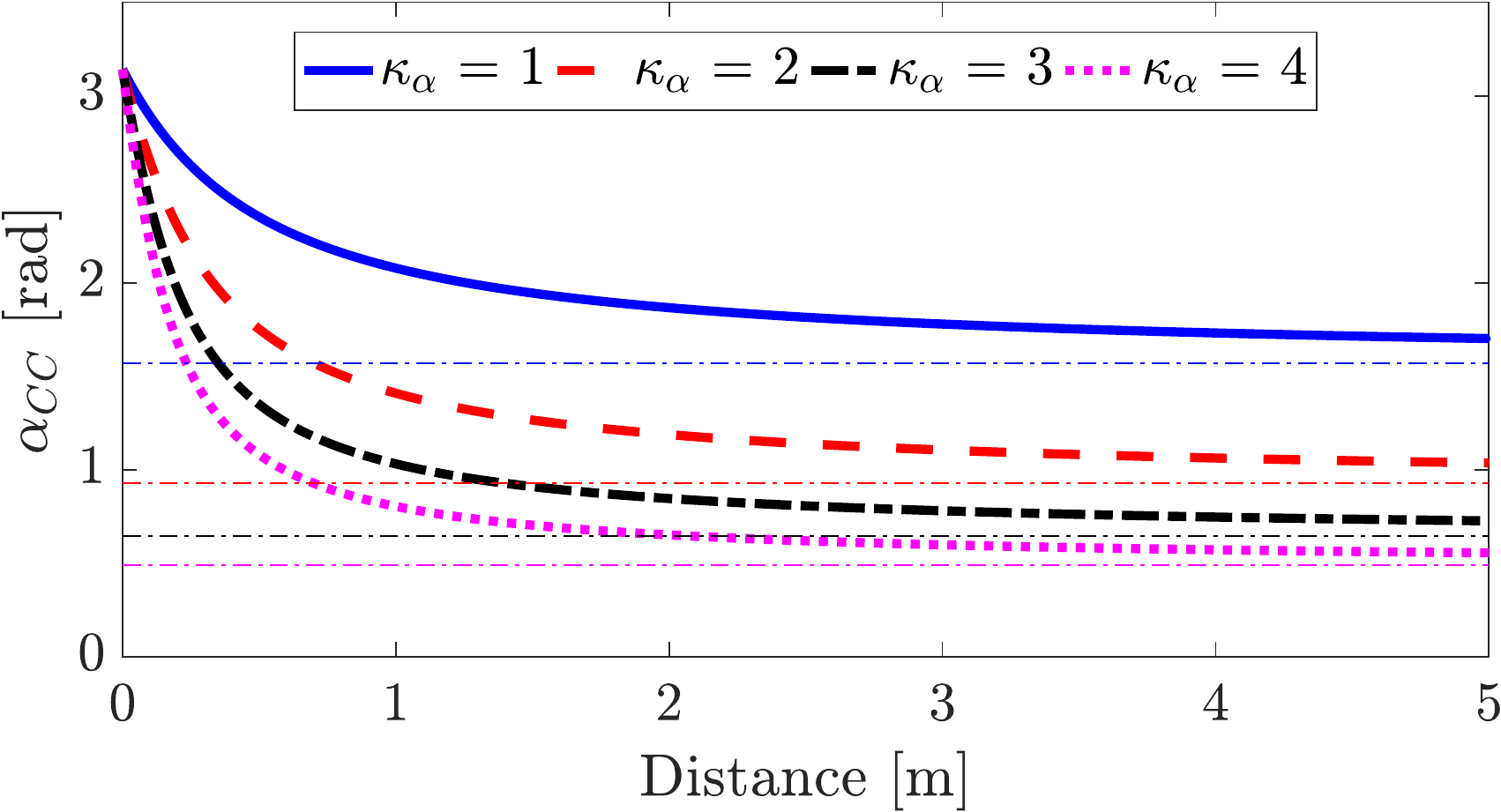}
    \caption{Effect of $\kappa_\alpha$ on $\alpha_{{CC}_{asymptote}}$ ($r = 0.1m$, $\varepsilon = 0.5$)}
    \label{fig:collisioncone_kappaalpha}
  \end{figure}

  The expansion angle of a collision cone is dependent on
    the distance between the MAVs (the MAV radii becomes more significant as distance decreases), and 
    the relative estimation errors \citep{conroy20143}.
  Based on this knowledge, we formulated \eqref{eq:alphacc} to calculate the expansion angle:
  \begin{equation}
    \alpha_{{CC}_{ji}} = 
    2*\tan^{-1}{\left(\frac{2r + {\bar{\rho}}_{ji} + \varepsilon_\alpha}{\kappa_\alpha*{\bar{\rho}}_{ji}}\right)},
    \label{eq:alphacc}
  \end{equation}
  where:
  $r$ is the radius of a \gls{MAV} (modeled as a circle);
  $\bar{\rho}_{ji}$ is the estimated range between $\mathcal{R}_i$ and $\mathcal{R}_j$;
  $\varepsilon_\alpha$ is an additional margin, the properties of which are discussed in \secref{sec:preservingbehavior};
  and $\kappa_\alpha$ is a coefficient describing the quality of the estimate.
  The expansion has a lower bound $\alpha_{{CC}_{asymptote}}$ which is dependent on $\kappa_\alpha$:
  \begin{equation}
    \alpha_{{CC}_{asymptote}} =\lim_{\bar{\rho}_{ji} \to \infty} \alpha_{{CC}_{ji}} =  2*\tan^{-1}{\left(\frac{1}{\kappa_\alpha}\right)}.
    \label{eq:alphaasymptote}
  \end{equation}
  Its impact may be appreciated in \figref{fig:collisioncone_kappaalpha}.
  In this work, unless otherwise stated, we use $\kappa_\alpha=1$, leading to $\alpha_{{CC}_{asymptote}}=\frac{\pi}{2}$.
  This incorporates the expected bearing errors expected during flight based on our preliminary results.
  
\subsection{Preserving Behavior in Rooms of Different Size}
\label{sec:preservingbehavior}
\begin{figure}[t]
    \centering
    \includegraphics[width=84mm]{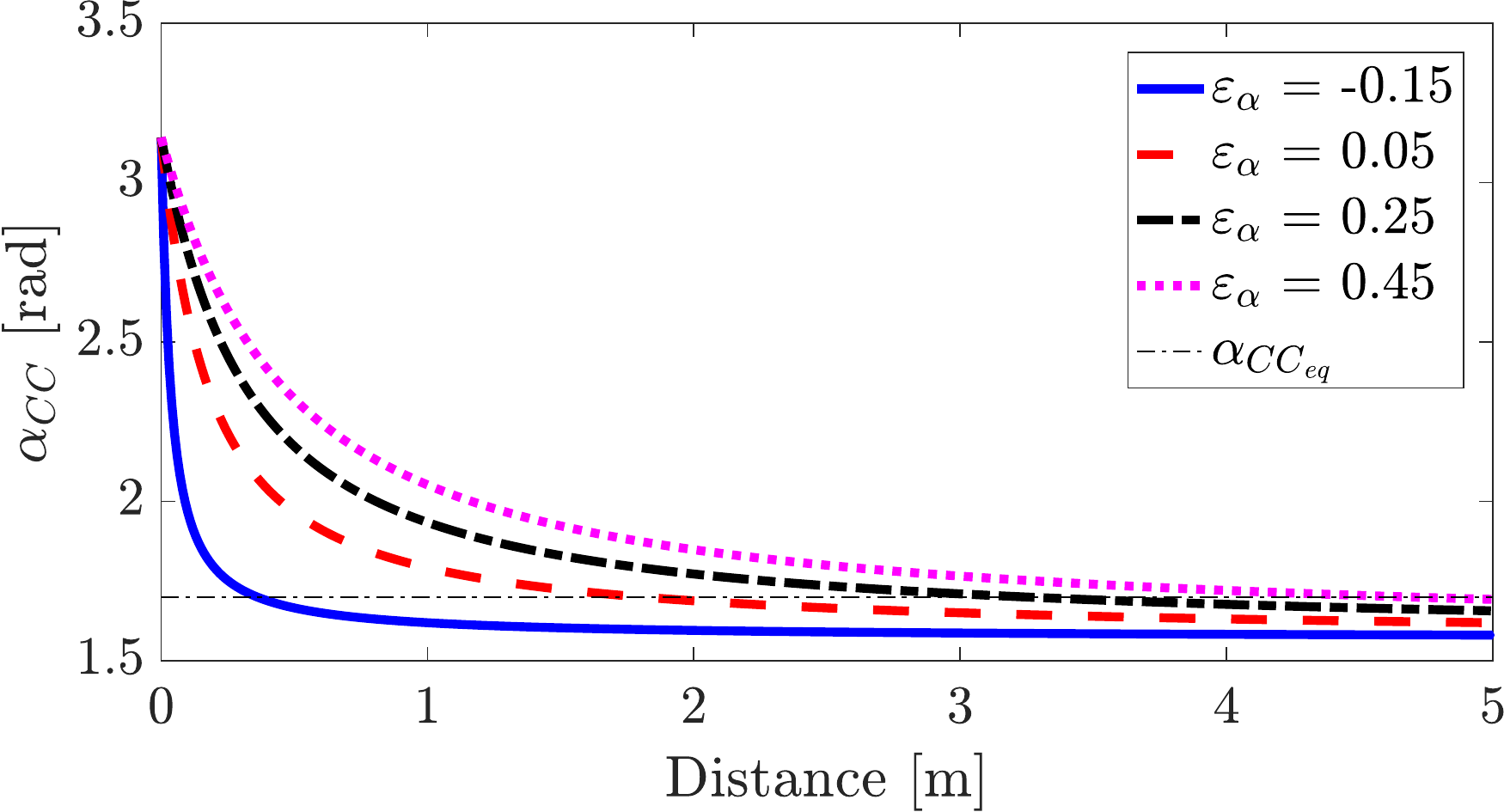}
    \caption{Effect of $\varepsilon_\alpha$ on $\alpha_{CC}$ ($r = 0.1m$, $\kappa_\alpha = 1$) }
    \label{fig:collisioncone_varepsilonalpha}
  \end{figure}

  The expansion angle of the collision cone widens towards $\pi$ as the distance between two MAVs decreases.
  This implies that in smaller rooms, the collision cones will \emph{always} feature wide expansion angles, leading to most of the environment becoming out of bounds.
  This restriction in freedom of movement creates oscillations in MAV trajectories.
  To solve the issue, we propose using the margin $\varepsilon_\alpha$ as a tuning parameter.
  The effect of varying $\varepsilon_\alpha$ may be appreciated with \figref{fig:collisioncone_varepsilonalpha}: 
  as $\varepsilon_\alpha$ decreases, the decay of the expansion angle with distance increases.
  A faster decay is suitable for smaller rooms so that motion is less restricted. \\

  We devised a method to tune $\varepsilon_\alpha$ intuitively.
  By re-arranging \eqnref{eq:alphacc}, $\varepsilon_\alpha$ is expressed by:
  \begin{equation}
    \varepsilon_\alpha = 
    \kappa_\alpha * \rho_{eq} * \tan{\left(\frac{\alpha_{{CC}_{eq}}}{2}\right)} - 2r - \rho_{eq},
    \label{eq:varepsilonalpha}
  \end{equation}
  This translates tuning $\varepsilon_\alpha$ to tuning a pair $\left\{\rho_{eq},\alpha_{{CC}_{eq}}\right\}$, where $\alpha_{{CC}_{eq}}$ is the desired angle of expansion at a distance $\rho_{eq}$.
  Note that $\alpha_{{CC}_{eq}} > \alpha_{{CC}_{asymptote}}$, and $\varepsilon_\alpha \geq -(r_i+r_j) $ if $\kappa_\alpha \geq 1 $.\\
  In all our tests, $\rho_{eq}$ is set to half of the side length of the room.
  $\alpha_{{CC}_{eq}}$ is kept at $1.7rad$.

\section{General Testing Methodology}
\label{sec:testsetup}
\subsection{Description of Arbitrary Task for Performance Testing}
\label{sec:taskdescription}
\begin{figure}[t]
    \centering
      \includegraphics[width=39mm]{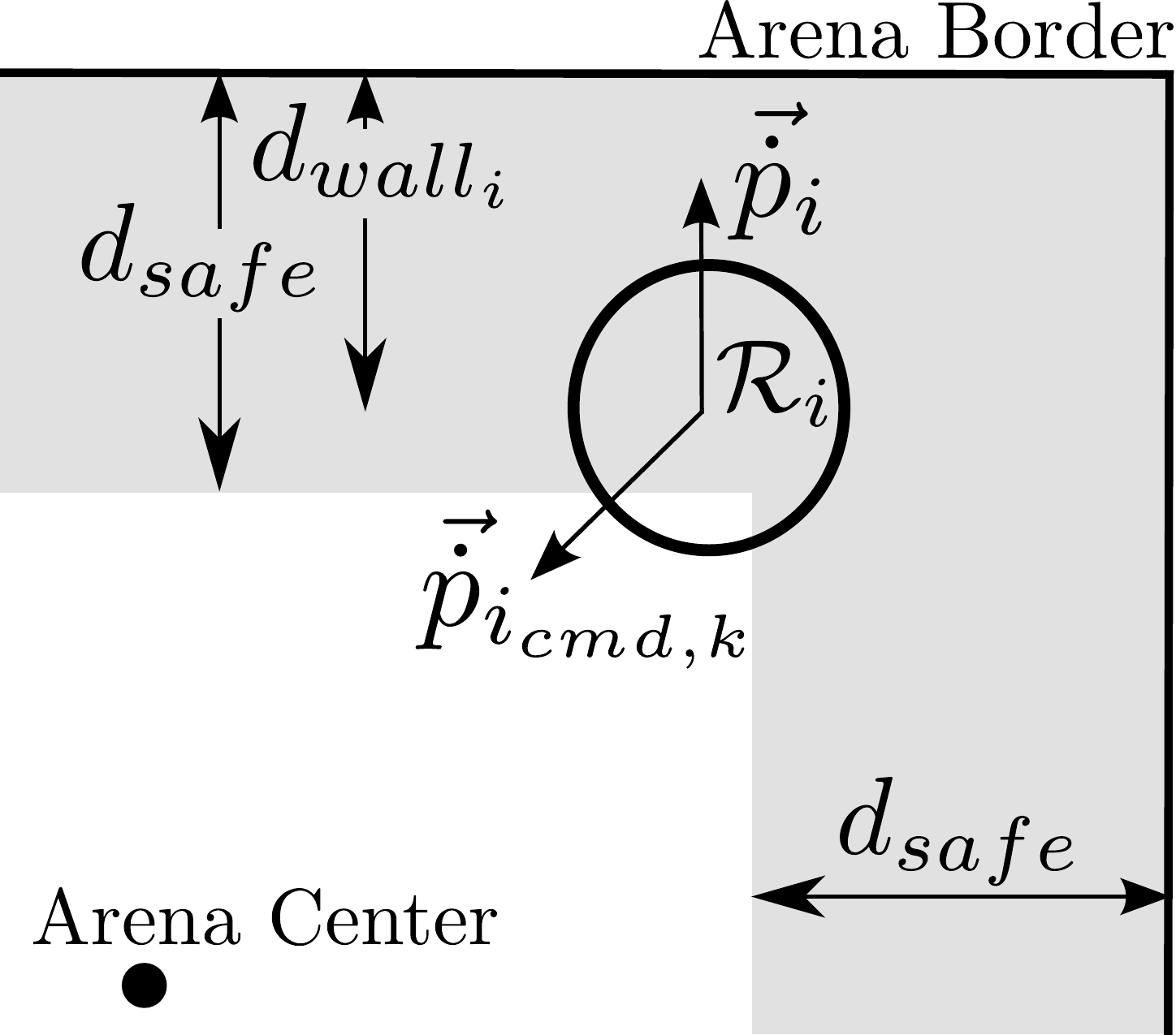}
      \caption{Depiction of condition M1}
      \label{fig:ConditionM1}
  \end{figure}

  A test exploration task was developed where multiple MAVs fly in a room, at the same altitude, and attempt to pass \emph{through} the center.
  This is designed to provoke collision and observe if/how they are resolved.
  Consider a team of $m$ homogeneous MAVs.
  Each MAV $\mathcal{R}_i$ can control its velocity.
  Let $\dot{\vec{p}}_{i_{cmd,k}}$ be the desired velocity for $\mathcal{R}_i$ expressed in its body-frame $\mathcal{F}_{B_i}$ at a given time-step $k$.
  Let $d_{{wall}_{i}}$ be the distance between $\mathcal{R}_i$ and the arena border that is closest to it, with $d_{safe}$ being a safety distance to the arena's borders.
  Remember that each robot $\mathcal{R}_i$ features $m-1$ EKF instances to keep track of the other members and uses their outputs to determine its collision cone set $CC_i$, see \eqnref{eq:ccsuperposition}.
  At each-time step $k$, the EKF outputs are updated and $CC_i$ is re-calculated.
  $\dot{\vec{p}}_{i_{cmd,k}}$ is then chosen as follows:
  $\dot{\vec{p}}_{i_{cmd,k}} = \dot{\vec{p}}_{i_{cmd,k-1}}$ unless conditions M1 and M2 take place.
  \renewcommand{\labelenumi}{M\theenumi:}
  \begin{enumerate}
    \item $d_{{wall}_{i}} < d_{safe}$ and $\dot{d}_{{wall}_{i}} < 0$.
    This means that $\mathcal{R}_i$ is close to the arena border and approaching it.
    Then, $\dot{\vec{p}}_{i_{cmd,k}}$ is rotated towards the center of the arena.
    See \figref{fig:ConditionM1}.
    \item $\dot{\vec{p}}_i \in CC_i$.
    This means that the current velocity of $\mathcal{R}_i$ could lead to a collision with one or more team members.
    An escape velocity is sought according to the strategy proposed in \secref{sec:avoidancebehavior}.
  \end{enumerate}
  \renewcommand{\labelenumi}{\theenumi.} 
  Condition M1 holds priority over M2 to ensure that the MAVs remain within the arena.
  At all time-steps, unless other-wise commanded by the collision avoidance algorithm, $\abs{\dot{\vec{p}}_{i_{cmd,k}}} = v_{nominal}$, where $v_{nominal}$ is a fixed speed magnitude.
  
\subsection{Assessment Strategy}
\label{sec:assessmentparameters}
  \paragraph {Assessment of Relative Localization: }
  During the task, this can be assessed by comparing the estimated relative locations to ground-truth data.

  \paragraph {Assessment of Collision Avoidance: }
  This is partially dependent on the performance of the relative localization, 
  yet can be assessed independently by identifying failure cases and observing general behavior properties.
  It is also interesting to determine the likelihood that the error falls within the expected collision cone bounds.

  \paragraph {Assessment of the Full System: } 
  The parameter of interest is the mean flight-time between collision.
  Ideally, a successful system is one that systematically ensures that collision will not take place.
  This metric is dependent on how crowded the airspace is.  
  By modeling MAVs as circles, airspace density is calculated with:
  \begin{equation}
    \mathcal{D}_{m,c} = \frac{m * \pi r_c^2}{s_c^2}
    \label{eq:density}
  \end{equation}
  $\mathcal{D}_{m,c}$ denotes the density for configuration $c$ with $m$ MAVs.
  $r_c$ is the radius of a MAV in configuration $c$.
  $s_c$ is the side length of the squared arena at configuration $c$.

  Experiments were performed in separate stages with increasing realism and autonomy, starting with simulation and ending with autonomously controlled flight with on-board measurements.
  After this, the technology was also ported and tested on miniature drones.
  The results of all tests are discussed in the next four chapters.
  Videos of the experiments are available at: \url{https://www.youtube.com/playlist?list=PL_KSX9GOn2P9f0qyWQNBMj7xpe1HARSpc}.

\section{Simulation Experiments}
\label{sec:simulationexperiments}
    Simulations allow to assess the collision avoidance algorithm and the full system.
    We can easily assess the performance of the system for several airspace densities, noise scenarios, etc., and obtain statistically relevant insights.

\subsection{Simulation Environment Set-Up}
\label{sec:simulationsetup}
\begin{figure}[t!]
  \centering
    \begin{subfigure}[t]{39mm}
      \centering
      \includegraphics[width=\textwidth]{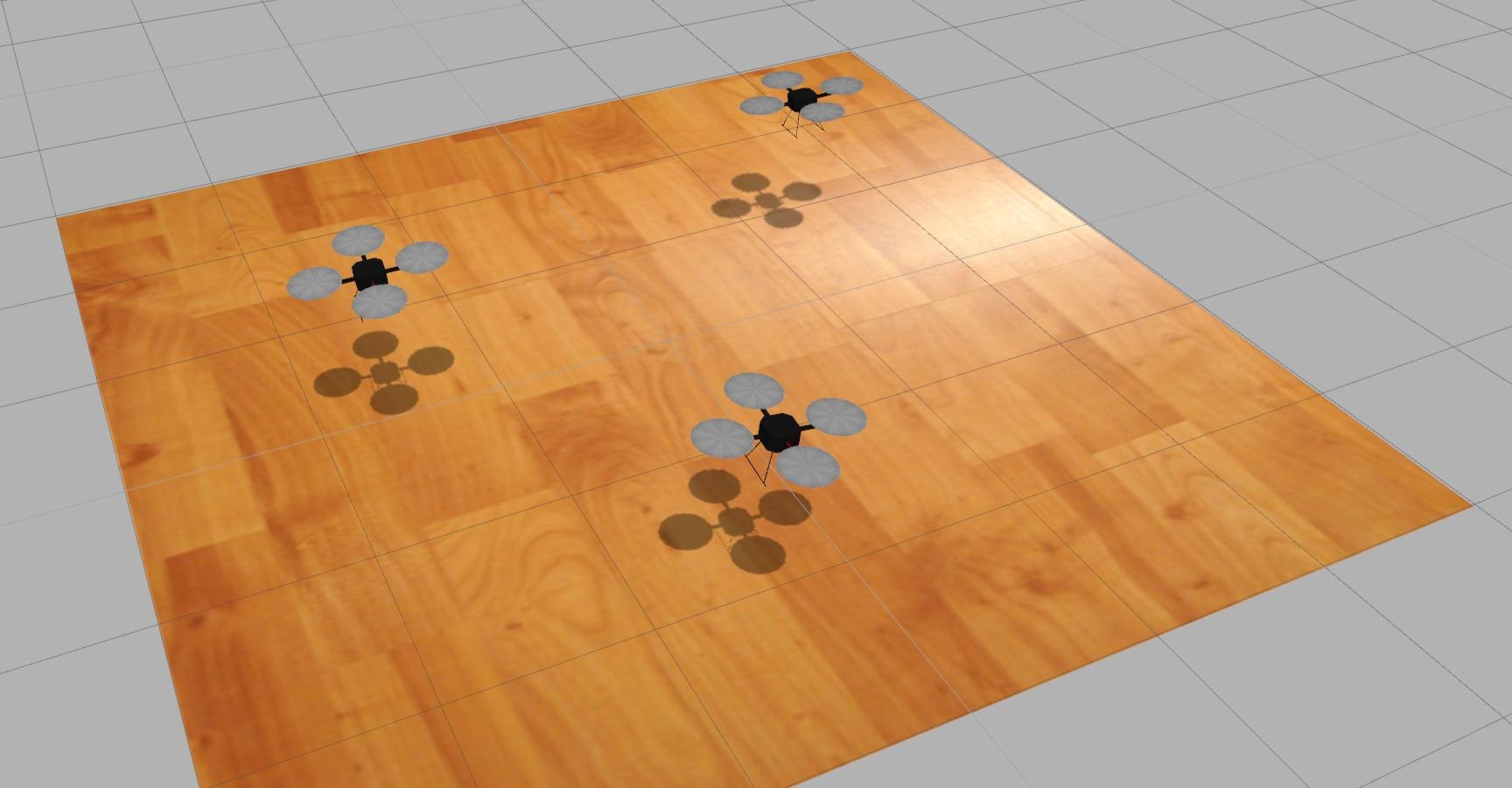}
      \caption{Screen-shot of a simulation with 3 MAVs}
      \label{fig:simulatedarena}
    \end{subfigure}
    ~
    \begin{subfigure}[t]{39mm}
      \centering
      \includegraphics[width=\textwidth]{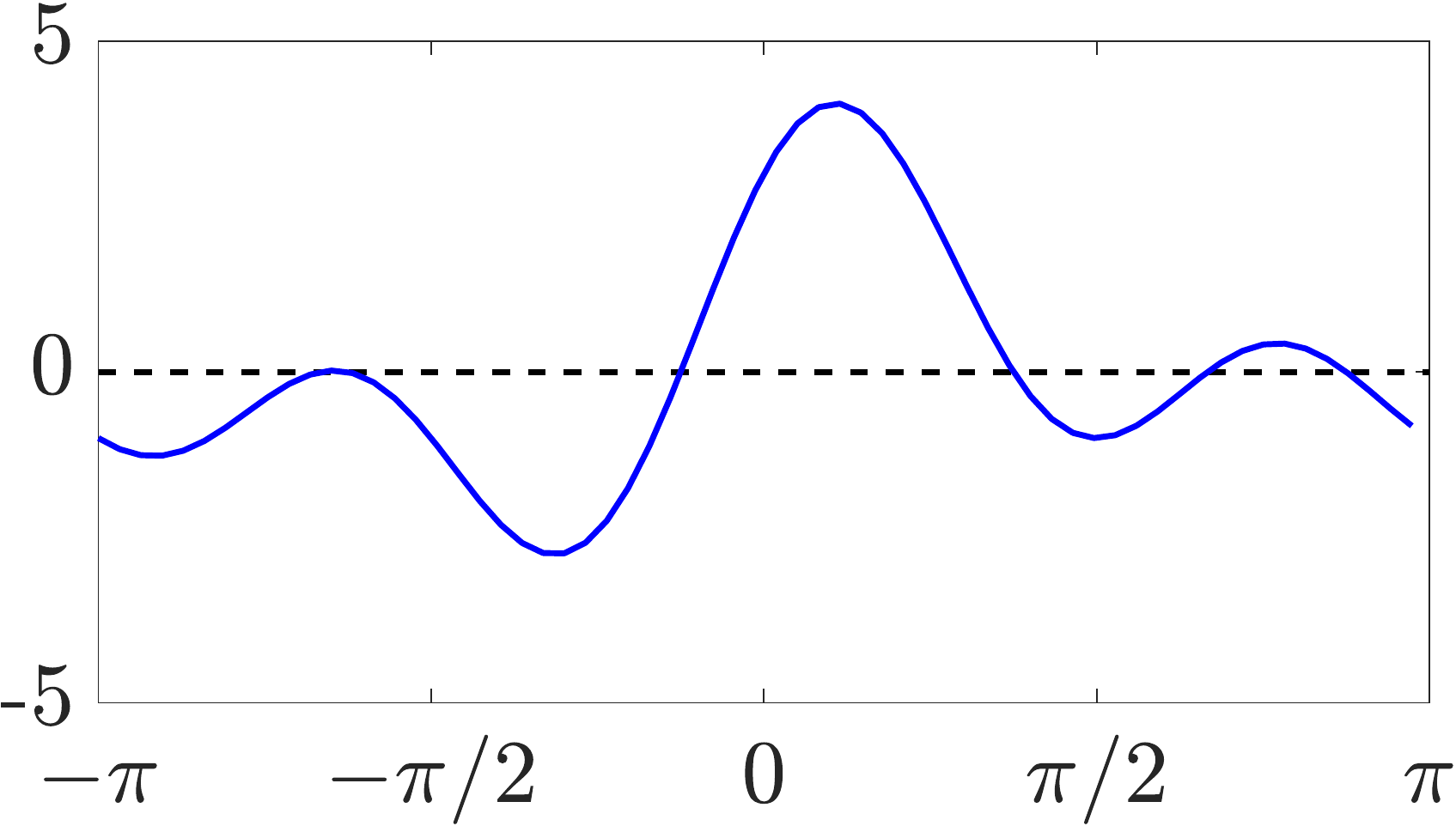}
      \caption{Simulated RSSI horizontal lobes applied as a function of relative bearing between MAVs}
      \label{fig:simulationassessment_lobes}
    \end{subfigure}
    \caption{Figures relating to the development of the simulation environment}
    \label{fig:simulationassessment}
  \end{figure}

  The simulation environment was developed using \gls{ROS} \citep{quigley2009ros}, 
  the \emph{Gazebo} physics engine \citep{koenig2004design}
  and the \emph{hector-quadrotor} model \citep{meyer2012comprehensive}.
  Multiple quad-rotor MAVs can be simulated simultaneously.
  A ROS module
  (or \emph{``node''})
  for each MAV simulates the Bluetooth communication and enforces the controller described in \secref{sec:taskdescription}.
  A rendered screen-shot of a simulation run is shown in \figref{fig:simulatedarena}.\\
  
  The \gls{RSSI} is simulated using the LD model 
  ($P_n = -63dB$, $\gamma_l=2.0$) 
  with added Gaussian noise (\gls{SD} of $5db$) and horizontal antenna lobes, unless otherwise stated.
  The lobes were modeled using a third order Fourier series with unitary weights, see \figref{fig:simulationassessment_lobes}.
  The other measurements were altered with the same standard deviations as in the preliminary localization tests of \secref{sec:localizationperformance_preliminary}.
  Furthermore: $v_{nominal} = 0.5m/s$, $d_{safe} = 0.25m$, and $\Psi = 0rad$ for all MAVs.
  The MAVs begin at different corners of the arena.
  The EKF is initialized such that the initial position guess is towards their initial flight direction (i.e. the center of the arena).\\
  
\begin{figure}[ht!]
    \centering
    \includegraphics[width=0.45\textwidth]{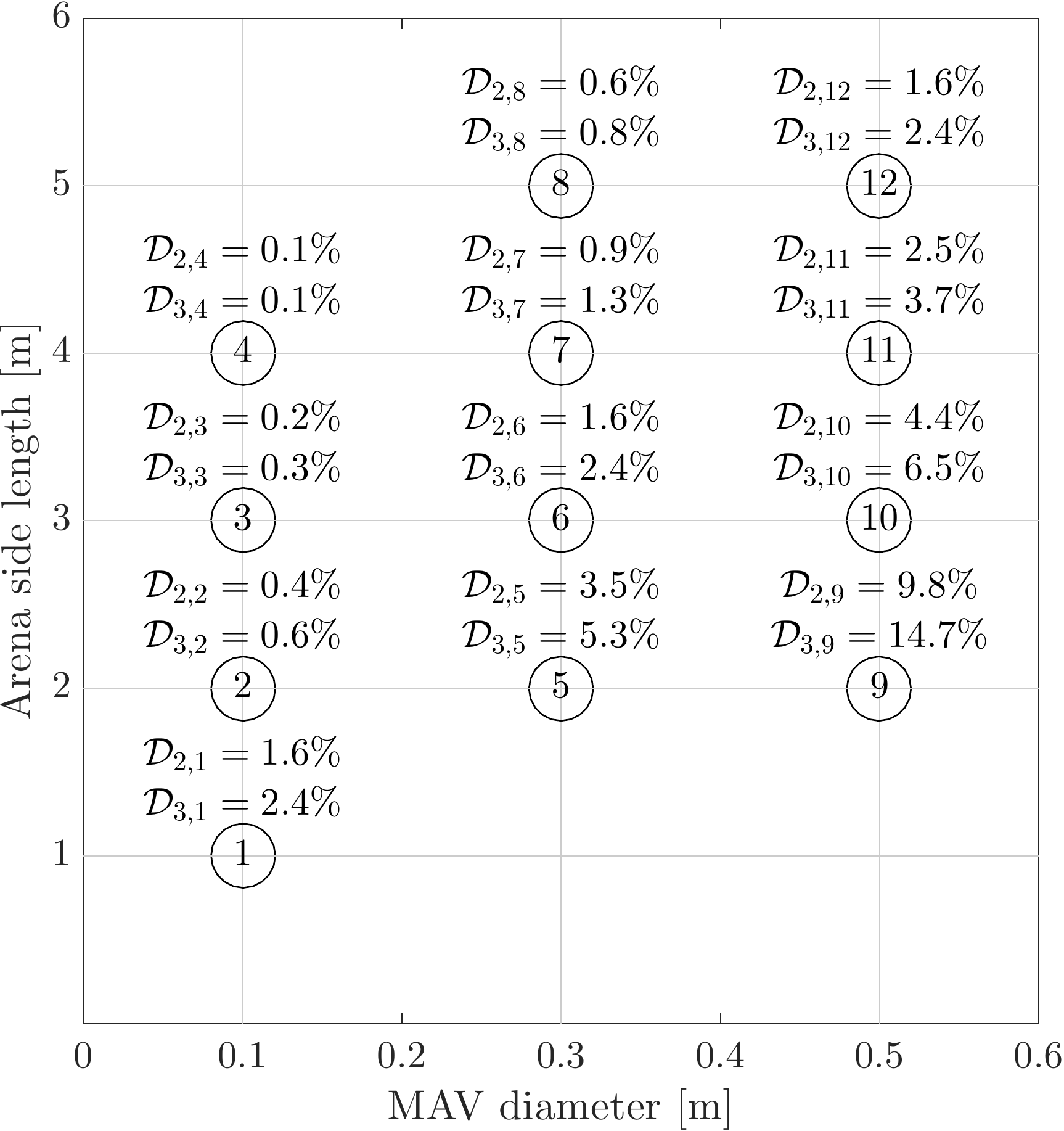}
    \caption{
    The twelve configurations tested in simulation with configuration numbers shown in the white circles}
    \label{fig:densitytable}
  \end{figure}

  We investigated twelve combinations of arena size and MAV diameter for teams of two MAVs and three MAVs.
  The combinations will be referred to by the encircled numbers in \figref{fig:densitytable}. 
  Each configuration was simulated 100 times.
  Each simulations was automatically interrupted if a collision occurred, or after $500s$ of collision-free flight.

\subsection{Results}
\label{sec:simulationresults}
\begin{figure}[t!]
    \centering
    \includegraphics[width=84mm]{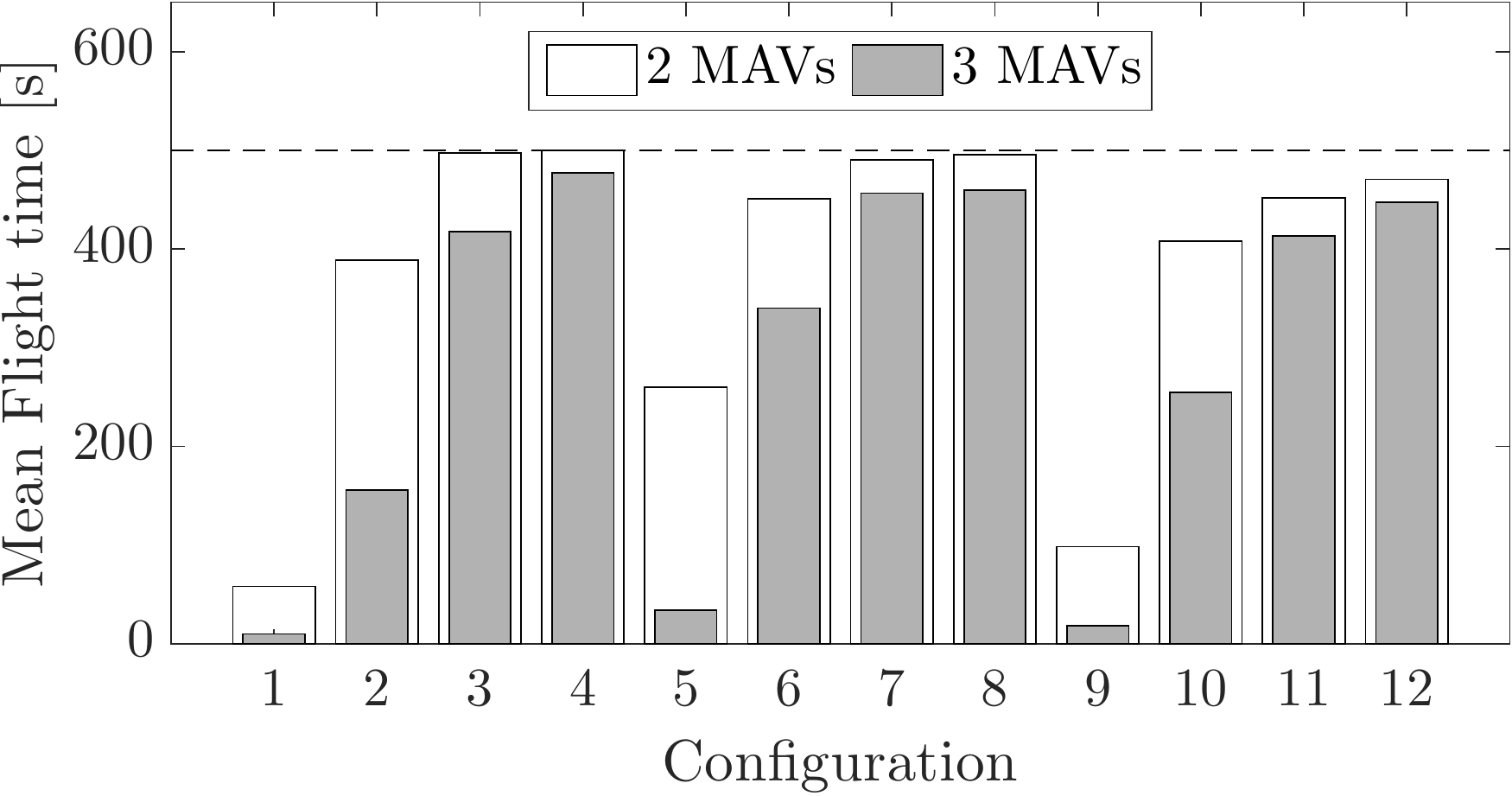}
    \caption{
      Mean flight-time to collision for all simulated configurations.
      Average results without collision avoidance, not shown in this figure, range between $3.9s$ and $14.3s$.
    }
    \label{fig:avoidanceresults_FT_bargraphs}
   \end{figure}

  Mean time-of-flight for each configuration is shown in \figref{fig:avoidanceresults_FT_bargraphs}.
  Flights with three MAVs consistently show a lower performance than with two MAVs.
  The performance drop is a result of the team dynamics at play, namely:
  \begin{inparaenum}[1)]
  \item increased airspace density; 
  \item decreased freedom of movement due to superposition of collision cones.
  \end{inparaenum}
  These two factors are analyzed in the remainder of this section. \\

   \begin{figure}[t!]
     \centering
     \begin{subfigure}[t]{39mm}
       \centering
       \includegraphics[width=\textwidth]{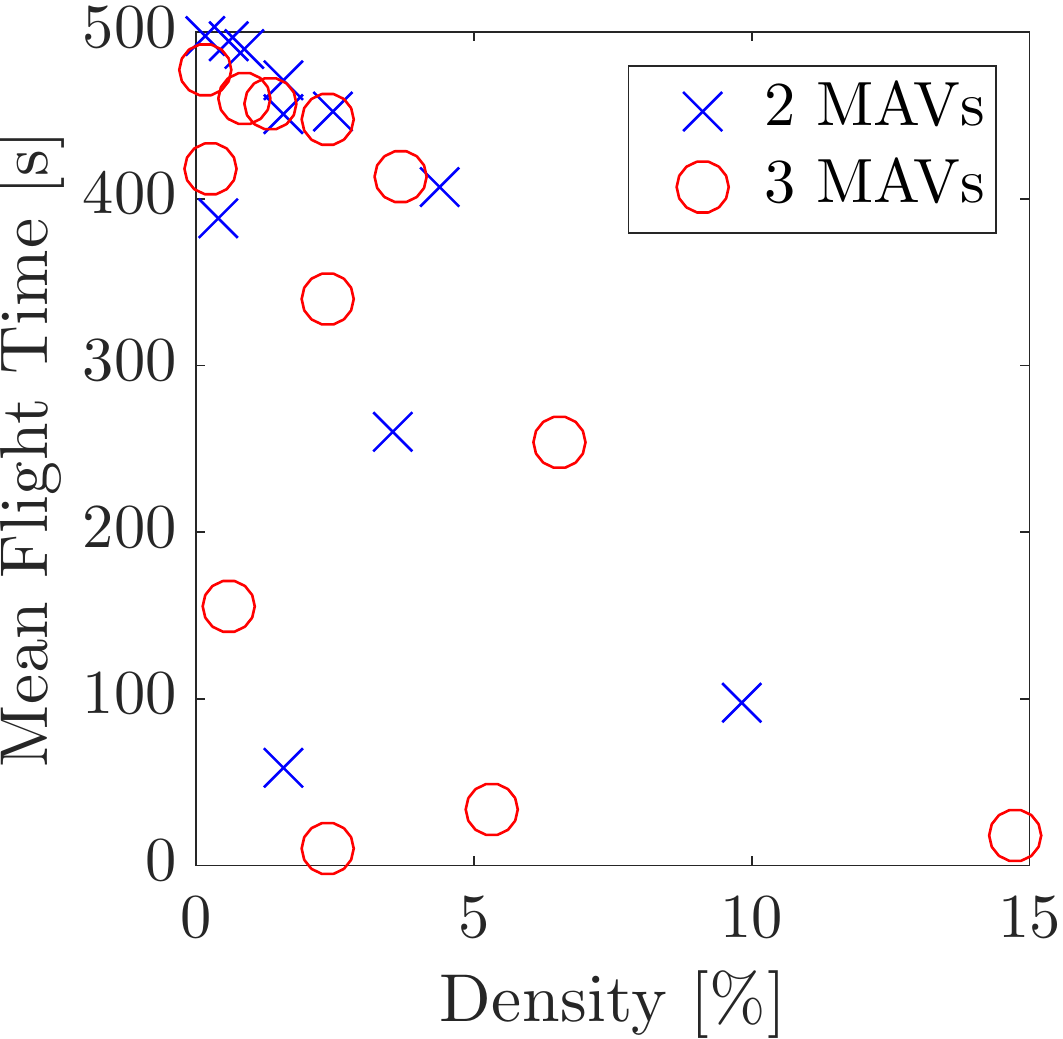}
       \caption{Mean flight-time with respect to density}
       \label{fig:flighttimedensity}
     \end{subfigure} 
     ~
     \begin{subfigure}[t]{39mm}
       \centering
       \includegraphics[width=\textwidth]{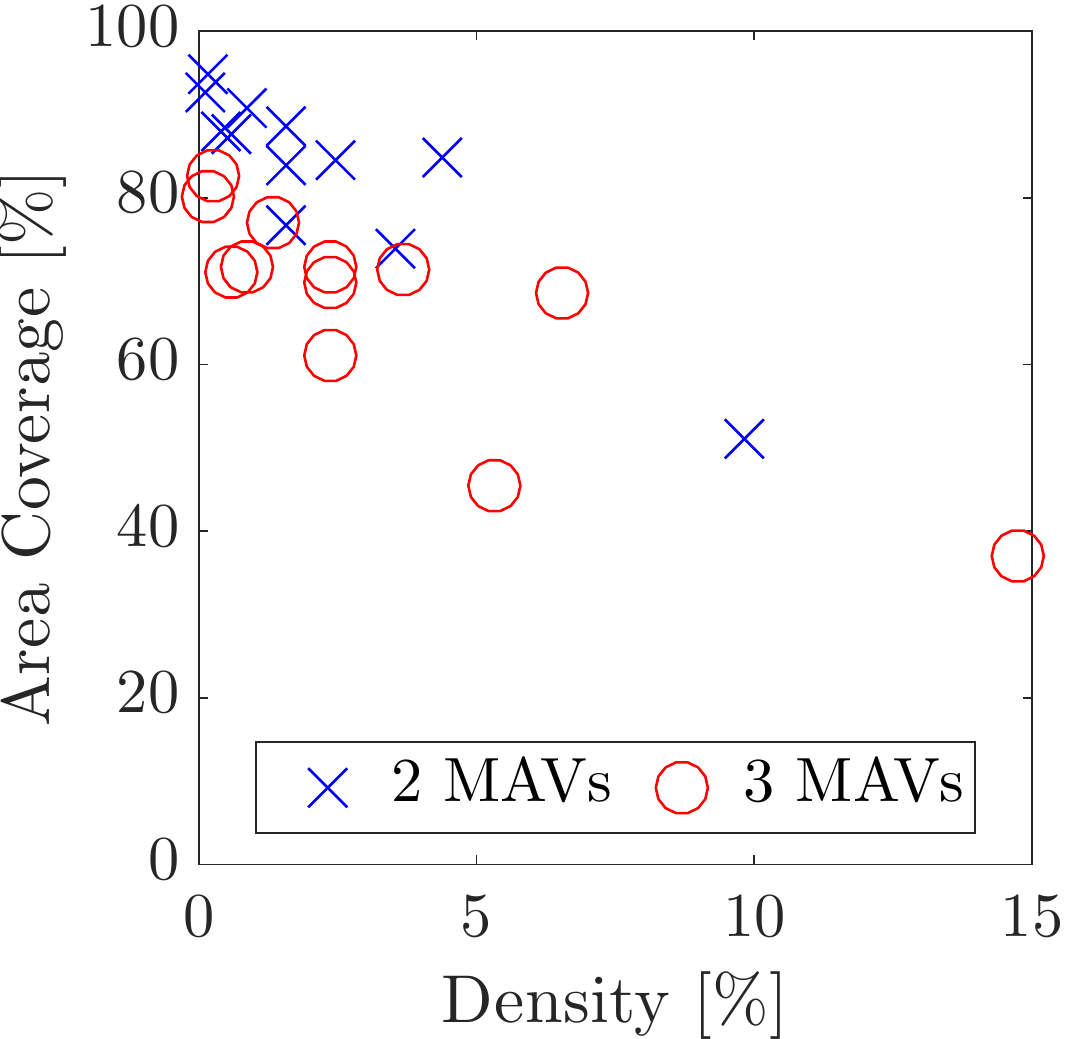}
       \caption{Mean area-coverage with respect to density}
       \label{fig:AreaCoverageDensity}
     \end{subfigure} 
     \caption{Flight parameters with respect to airspace density based on simulation results}
     \label{fig:simulationdensityrelations}
   \end{figure}
   When the arena side-length remains constant and the MAV diameter increases, a decrease in mean flight-time is observed.
   This is seen by comparing within the configuration triads 4-7-11, 3-6-10, and 2-5-9, and the pair 8-12.
   The result is analogous when MAVs of the same diameter are used in arenas of different sizes, see the configuration quartets 1-2-3-4, 5-6-7-8, and 9-10-11-12.
   This implies that a lower density improves the probability of success, but this is found to not strictly be the case.
   \figref{fig:flighttimedensity} shows the flight time to collision as a function of the airspace density.
   A portion of configurations show low results in spite of the low airspace density, and are outliers in the negative linear trend.
   These correspond to configurations 1, 2, 5, and 9, which feature smaller arena sizes.
   The conclusion is that room size affects performance even when airspace density remains constant.
   This is a remaining limitation of the current status of the system when operating in smaller room sizes.
   Its causes are discussed in \secref{sec:discussion_localizationperformance}.\\

\begin{figure}[t!]
    \centering
    \begin{subfigure}[t]{39mm}
    \centering
    \includegraphics[width=\textwidth]{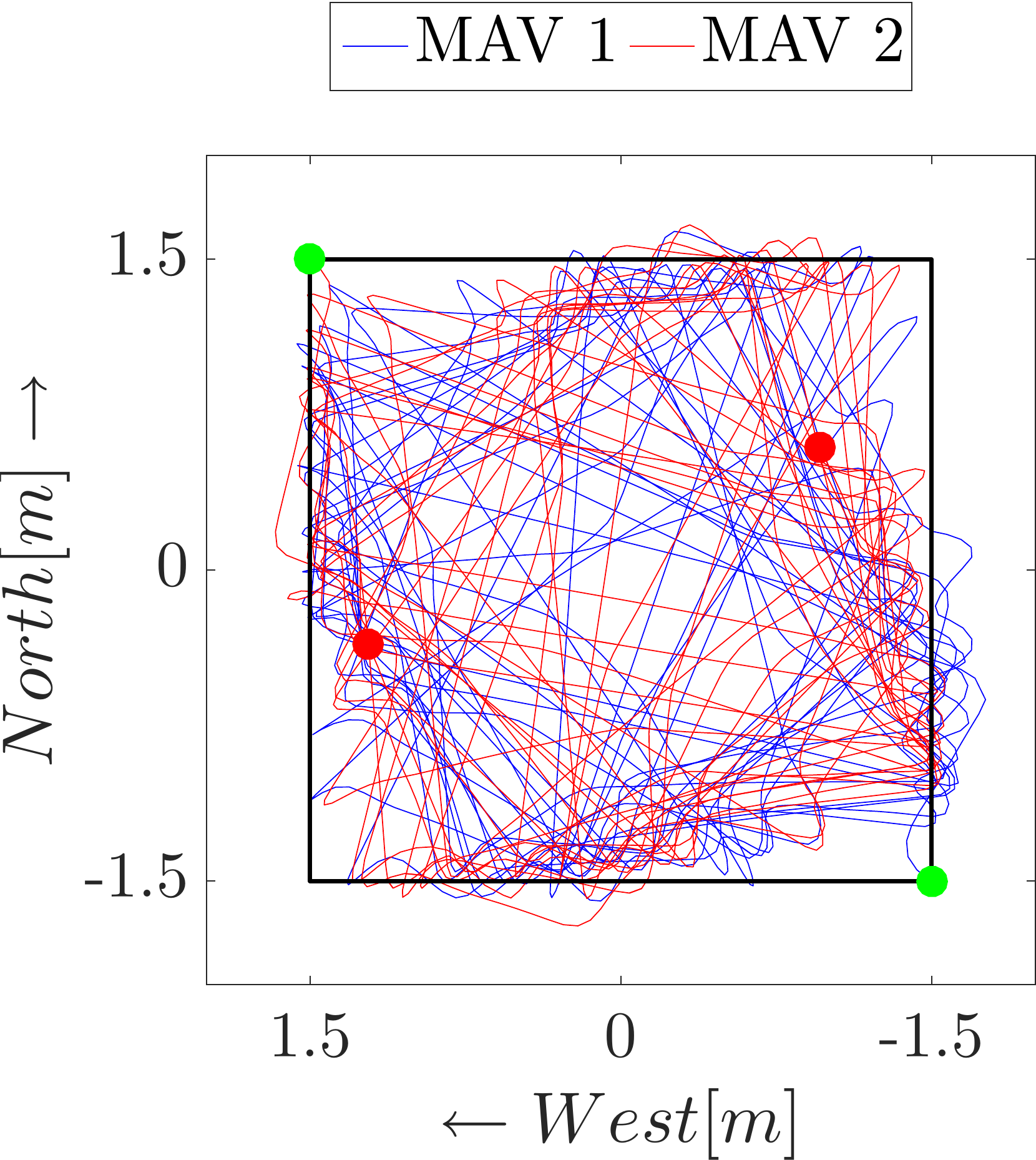}
    \caption{Two MAVs}
    \label{fig:behavior_circle_2}
    \end{subfigure} 
    ~
    \begin{subfigure}[t]{39mm}
      \centering
      \includegraphics[width=\textwidth]{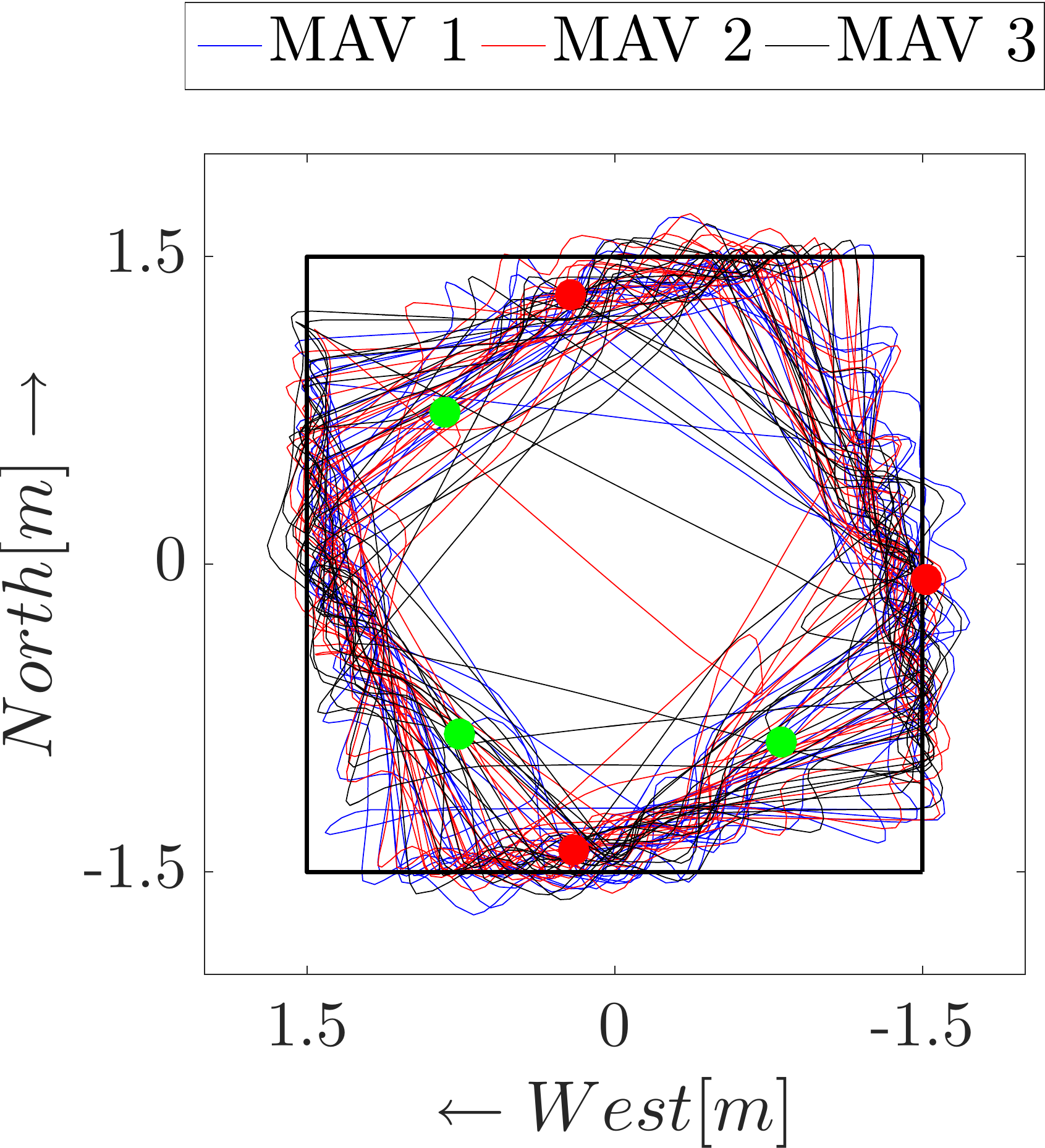}
      \caption{Three MAVs}
      \label{fig:behavior_circle_3}
    \end{subfigure} 
    \caption{Emergent circular behavior from two exemplary flights of $500s$ extracted from configuration 10 (green = starting position, red = final position)}
    \label{fig:behavior_circle}
  \end{figure}

   \figref{fig:AreaCoverageDensity} shows the impact of airspace density on area coverage for all flights with two MAVs and three MAVs.
   Area coverage was measured as follows.
   The total area is divided in sections of $0.20m\times0.20m$.
   A section is marked ``covered'' when one of the MAVs crosses it during a trial.
   Area coverage is the percentage of covered sections.
   With this, two patterns arise.
   \begin{enumerate*}
    \item A higher airspace density leads to a lower overall coverage.
    This is due to:
      \begin{inparaenum}[a)] 
        \item lower flight times, providing less overall time to complete the mission, and 
        \item decreased freedom of movement due to larger portions of the arena being covered by collision cones.
      \end{inparaenum}
    \item Three MAVs systematically achieve lower area coverage than only two MAVs in the same configuration.
    This is explained by analyzing the flight trajectories in more detail, from which an emergent circular behavior is discerned.
    See, for instance, \figref{fig:behavior_circle}, which shows two exemplary runs from a simulation with two
    (\figref{fig:behavior_circle_2})
    and three
    (\figref{fig:behavior_circle_3})
    MAVs from configuration 10.
    When more than one MAV to avoid is present, the superposition of multiple collision cones significantly discourages the pursuit of the desired trajectory.
    The result is clock-wise motion along the sides of the arena for all MAVs.
    Oscillations along the border are observed as conditions M1 and M2 alternate.
   \end{enumerate*}

\subsection{Impact of RSSI Noise on Performance}
\label{sec:discussion_varyingnoiselevels}
\begin{figure}[t!]
    \centering
    \begin{subfigure}[t]{0.23\textwidth}
      \centering
      \includegraphics[width=\textwidth]{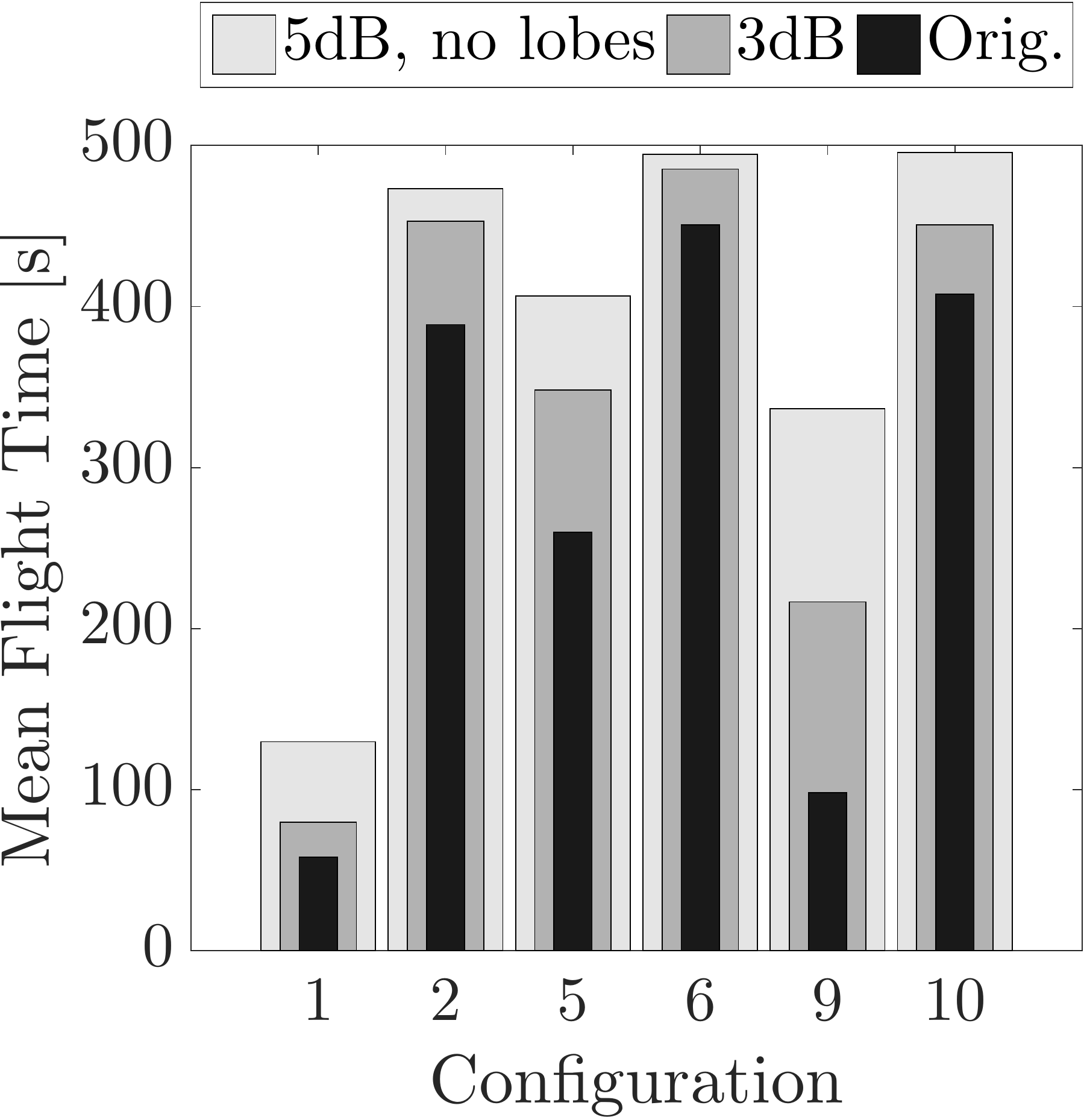}
      \caption{Two MAVs}
      \label{fig:noiseimprovement2}
    \end{subfigure} 
    ~
    \begin{subfigure}[t]{0.23\textwidth}
      \centering
      \includegraphics[width=\textwidth]{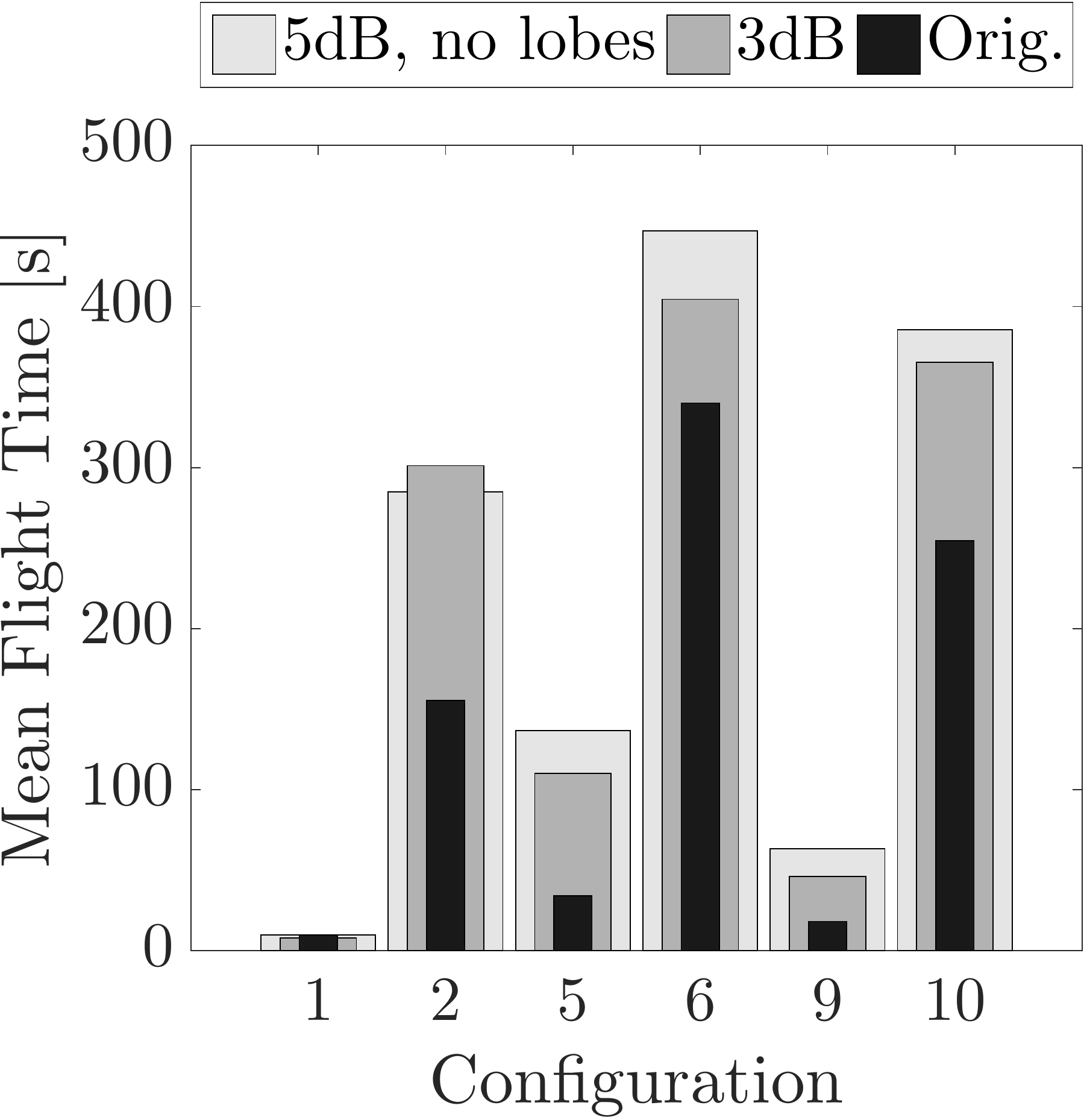}
      \caption{Three MAVs}
      \label{fig:noiseimprovement3}
    \end{subfigure} 
    \caption{Improvements in system performance against nominal results (``Orig.'', black, narrowest) when noise is reduced from $5dB$ to $3dB$ (dark gray, mid width) or when lobes are removed (dark gray, widest)}
    \label{fig:noiseimprovements}
  \end{figure}

  In simulation, two further case-studies were explored.
  In the first case, the simulated RSSI noise is reduced from $5dB$ to $3dB$, but lobes are still simulated.
  In the second case, RSSI noise is kept at $5dB$ but sensor lobes are removed.
  All other parameters remain the same as in the primary simulations.
  The configurations tested are those with the lowest performance: 
  1, 2, 5, 6, 9, 10.
  The results are shown in \figref{fig:noiseimprovements}, and show that removing the antenna lobes provides the largest improvement in performance.
  A lower noise also improves results, yet the impact is generally lower than antenna lobes.
  The lower error in relative position estimates translates to a more successful collision avoidance system.
  This implies that performance could be improved further if operating in cleaner environments, if using better antennas, or with a better filtering of noises.

\section{Experiments featuring External Own-State Measurements}
\label{sec:realworldexperiments_controlled_setup}
    These experiments use Optitrack to accurately inform MAVs of their velocity, orientation, and altitude. 
    This isolates the impact of using real RSSI measurements and Bluetooth communication on the relative localization system during flight.
    It also provides system performance data in the case of high-quality on-board estimates.
\subsection{Experimental Set-Up}
\label{sec:realworldsetup_controlled}

  \begin{figure}[t!]
    \centering
    \includegraphics[width=84mm]{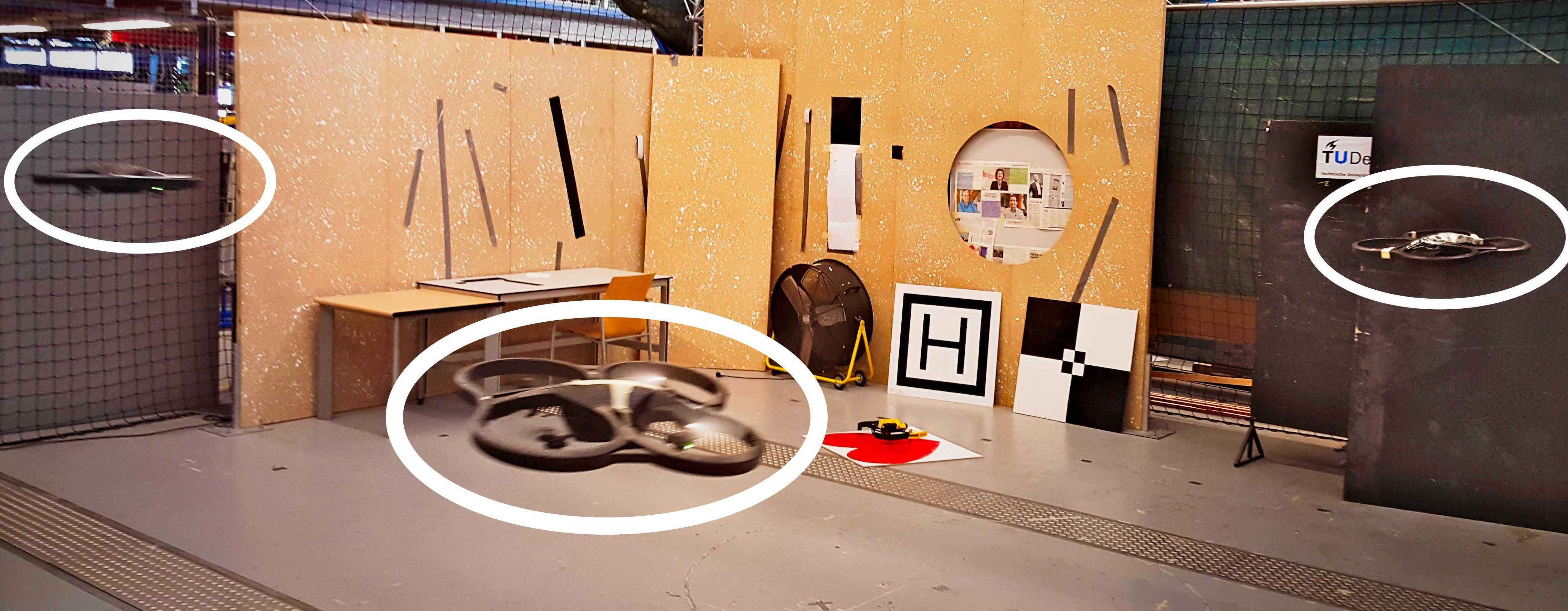}
    \caption{A flight with 3 AR.Drones (encircled in white)}
    \label{fig:experiment}
  \end{figure}
  \begin{figure*}[t]
    \centering
    \begin{subfigure}[t]{84mm}
      \centering
      \includegraphics[width=\textwidth]{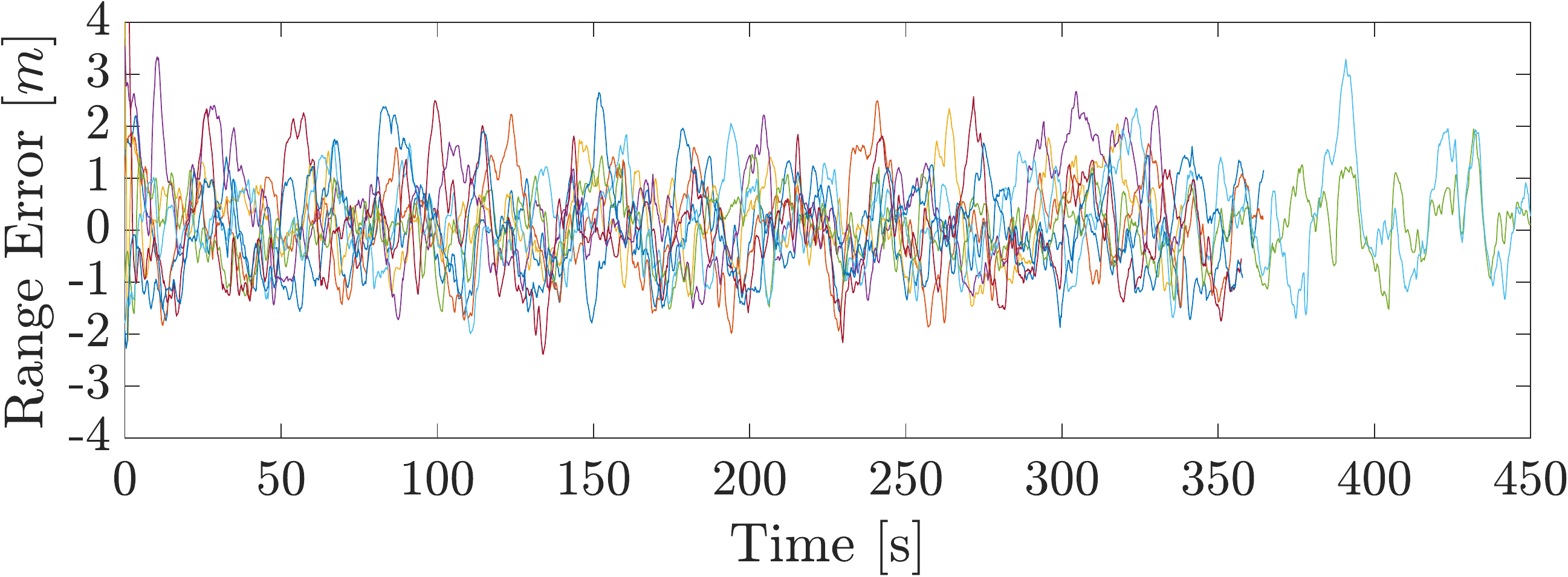}
      \caption{Range estimate error with two AR.Drones (RMSE=$0.86m$)}
      \label{fig:rangeerror_controlled_2mav}
    \end{subfigure} 
    ~
    \begin{subfigure}[t]{84mm}
      \centering
      \includegraphics[width=\textwidth]{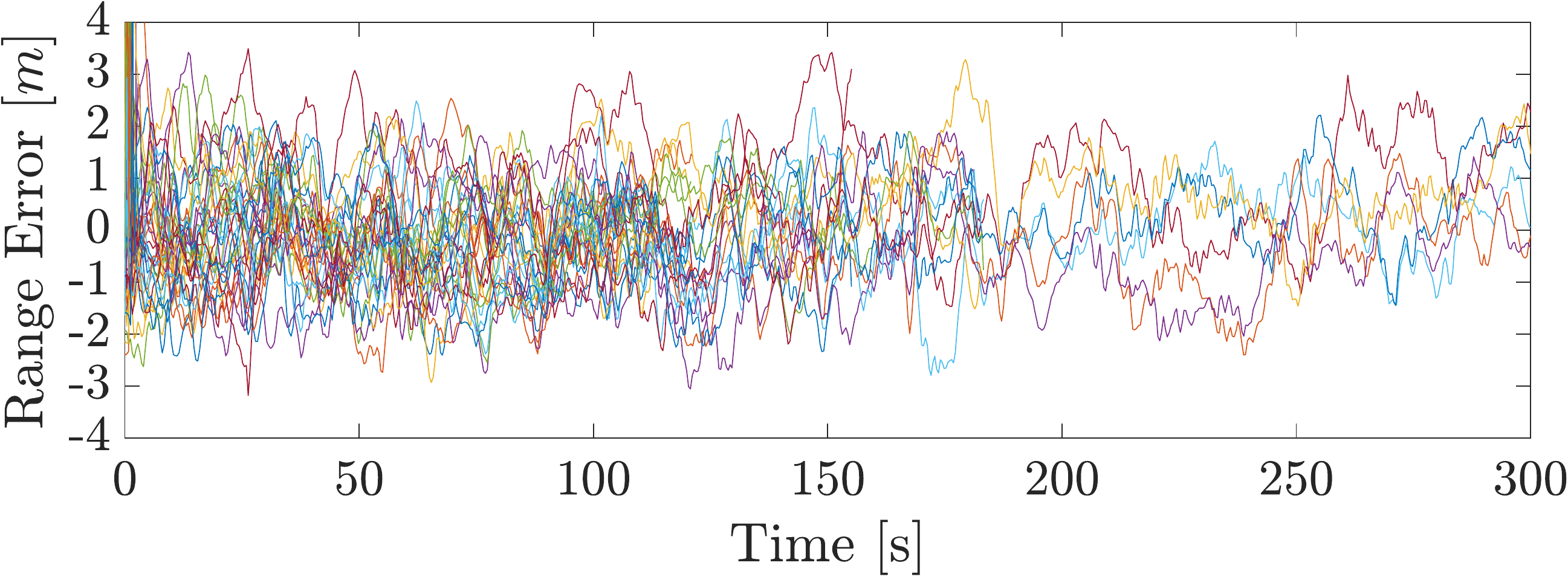}
      \caption{Range estimate error with three AR.Drones (RMSE=$1.14m$)}
      \label{fig:rangeerror_controlled_3mav}
    \end{subfigure}
    \begin{subfigure}[t]{84mm}
      \centering
      \includegraphics[width=\textwidth]{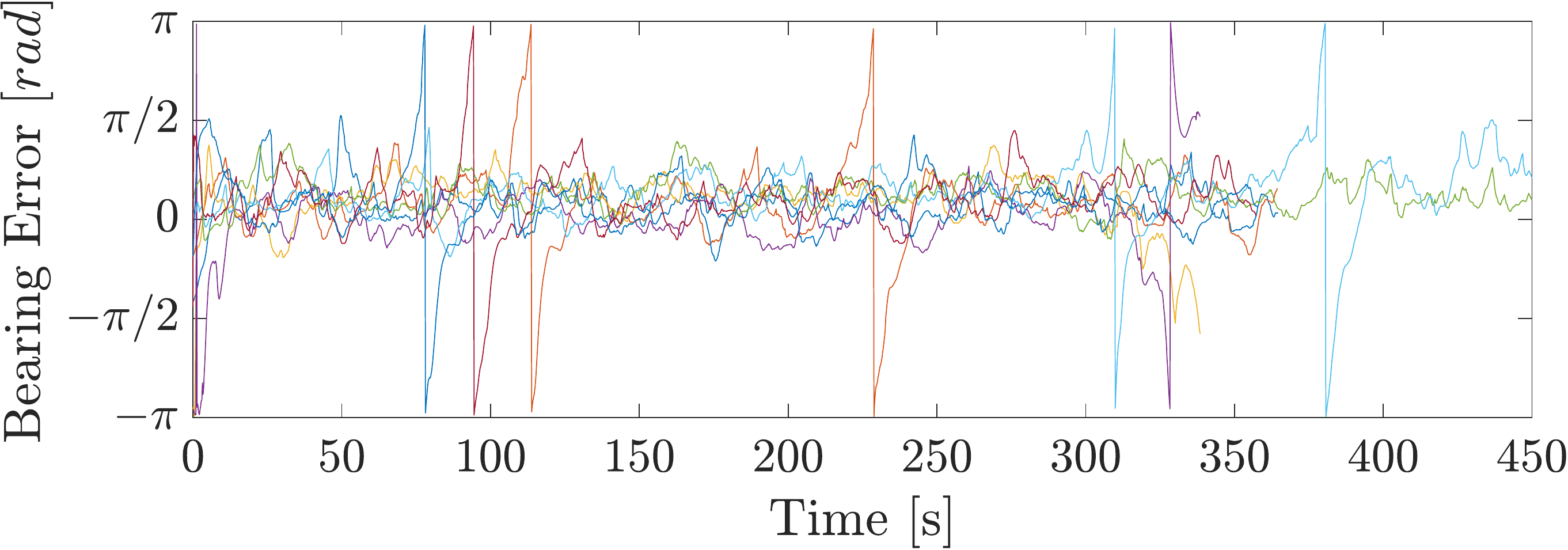}
      \caption{Bearing estimate error with two AR.Drones (RMSE=$0.57rad$)}
      \label{fig:bearingerror_controlled_2mav}
    \end{subfigure} 
    ~
    \begin{subfigure}[t]{84mm}
      \centering
      \includegraphics[width=\textwidth]{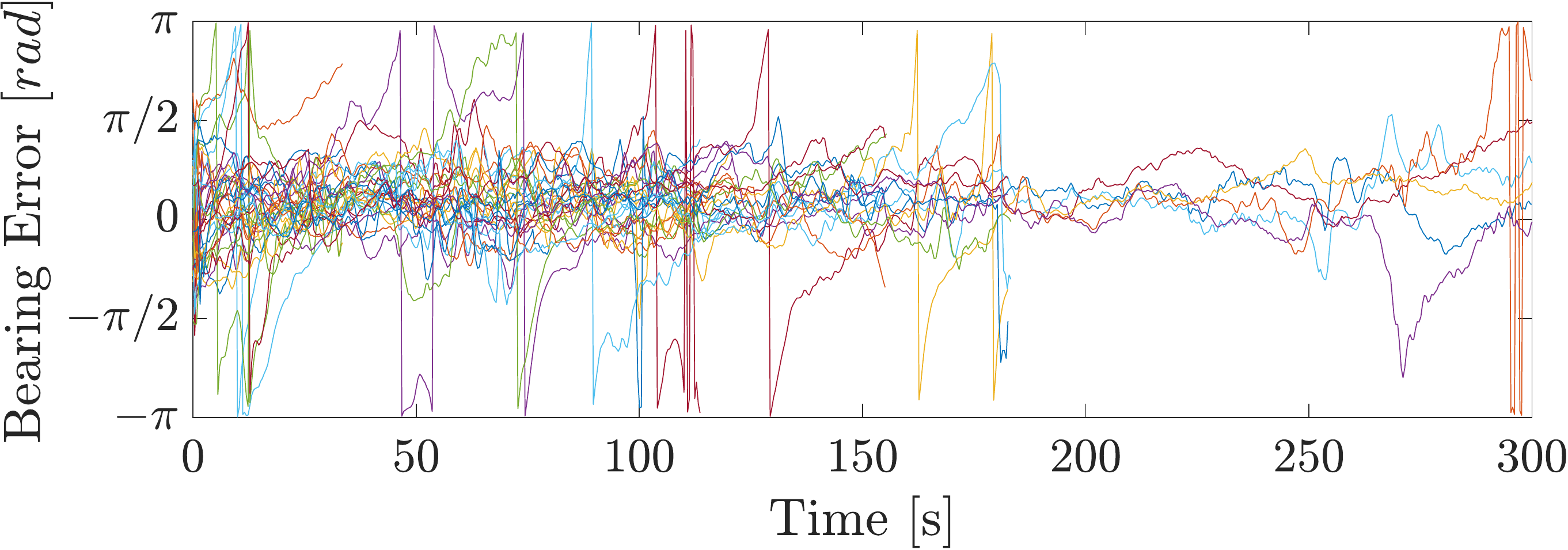}
      \caption{Bearing estimate error with three AR.Drones (RMSE=$0.70rad$)}
      \label{fig:bearingerror_controlled_3mav}
    \end{subfigure}
    \caption{Overview of all relative range (a,b) and relative bearing (c,d) estimation errors for flights with external state measurements}
    \label{fig:exp_errorplots}
  \end{figure*} 

  These experiments were performed using AR.Drones 2.0 \citep{parrot2012drone}.
  A BLED112 \citep{bleddongle} Bluetooth Smart USB Dongle enabled them with Bluetooth.
  The controller was developed using Paparazzi \citep{mueller2007paparazzi} and was running entirely on-board.
  The experiments in this section relied on Optitrack 
  to provide each MAV with data of its own velocity, orientation, and height via a Wi-Fi link.
  Each AR.Drone then communicated this data via the Bluetooth broadcast to the other ones. \\
  
  All MAVs flew at $1.5m$ from the ground, with a nominal speed $v_{nominal}=0.5m/s$ and safety wall distance $d_{safe}=0.5m$.
  The enforced arena size in all experiments was $4m\times4m$,
  making these tests analogous to Configuration 11 from the simulation runs
  (AR.Drones are slightly larger in diameter than $0.5m$).
  The LD model in the EKF filter was tuned with: $P_n = -68dB$ and $\gamma_l=2.0$.
  $P_n$ was obtained using a brief hand-held measurement, 
  $\gamma_l$ was based on the free-space assumption.
  The Optitrack measurements inputted into the EKFs were altered with Gaussian noises $\sigma_v = 0.2m/s$ and $\sigma_\psi = 0.2rad$.
  \figref{fig:experiment} shows a picture of a flight with 3 AR.Drones.

\subsection{Results}
\label{sec:realworldresults_controlled_results}
  Four flights were performed with two AR.Drones for a cumulative time of $25.3min$.
  Only one collision took place, which occurred in the second flight after $5.6min$.
  The other flights lasted $6.1min$, $7.6min$, and $6.0min$ without collisions;
  they were ended manually due to low battery.\\

  Six controlled flights were performed with three AR.Drones for a cumulative time of $15.3min$.
  Five flights ended in collisions.
  The flights ending with collisions reached featured a mean flight time of $160s$ ($2.7min$).
  The shortest flight was $33s$, the longest was $5.2min$.
  The other flights lasted $1.9min$, $2.6min$, and $3.0min$.
  The flight without a collision was manually ended after $2.0min$ due to low battery.
  Overall, this set-up with three MAVs can expect a collision once every $184s$ ($\approx3min$).\\
  \begin{figure}[t!]
    \centering
    \begin{subfigure}[t]{39mm}
      \centering
      \includegraphics[width=\textwidth]{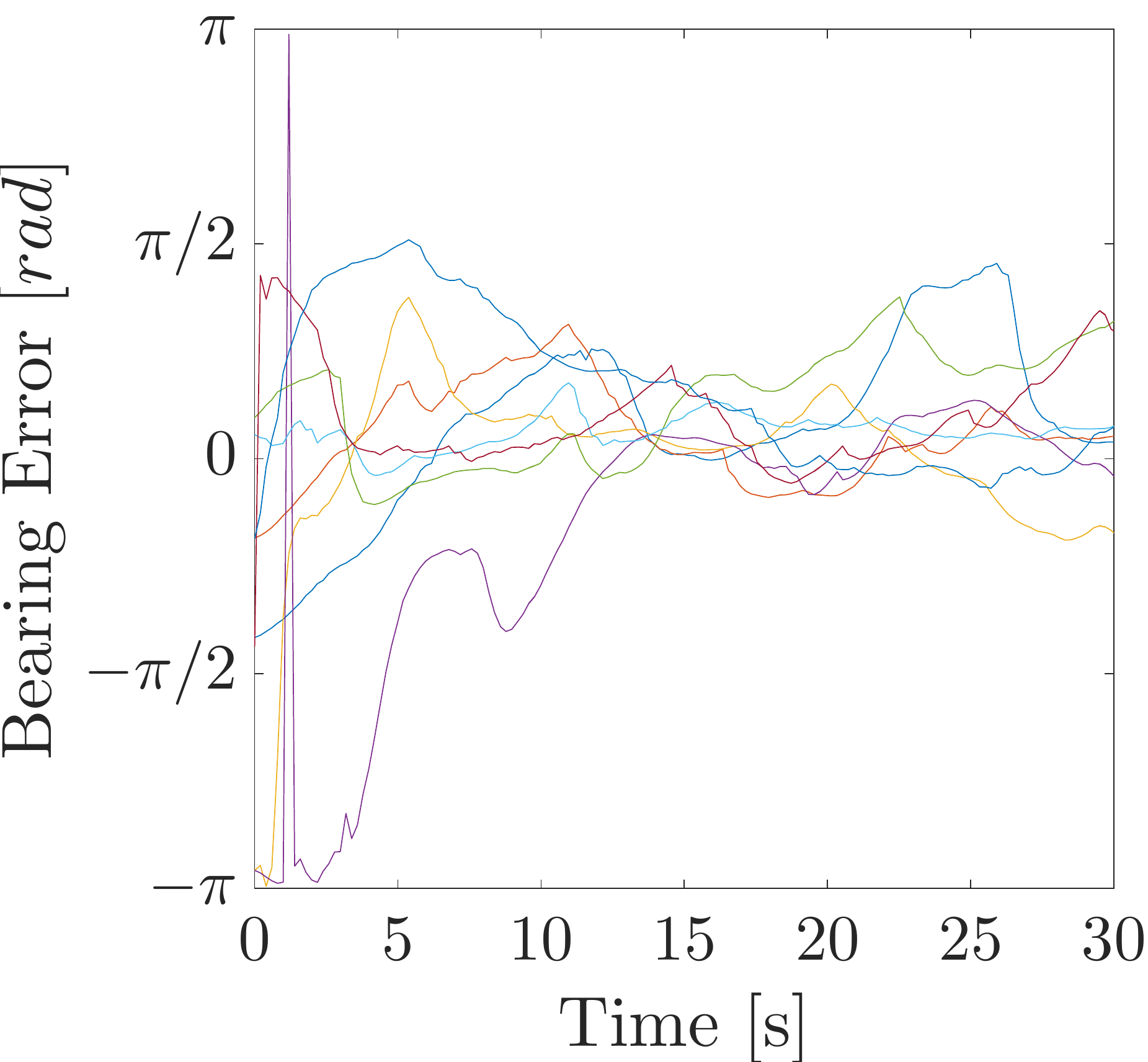}
      \caption{With two AR.Drones}
      \label{fig:error_convergence_2}
    \end{subfigure} 
    ~
    \begin{subfigure}[t]{39mm}
      \centering
      \includegraphics[width=\textwidth]{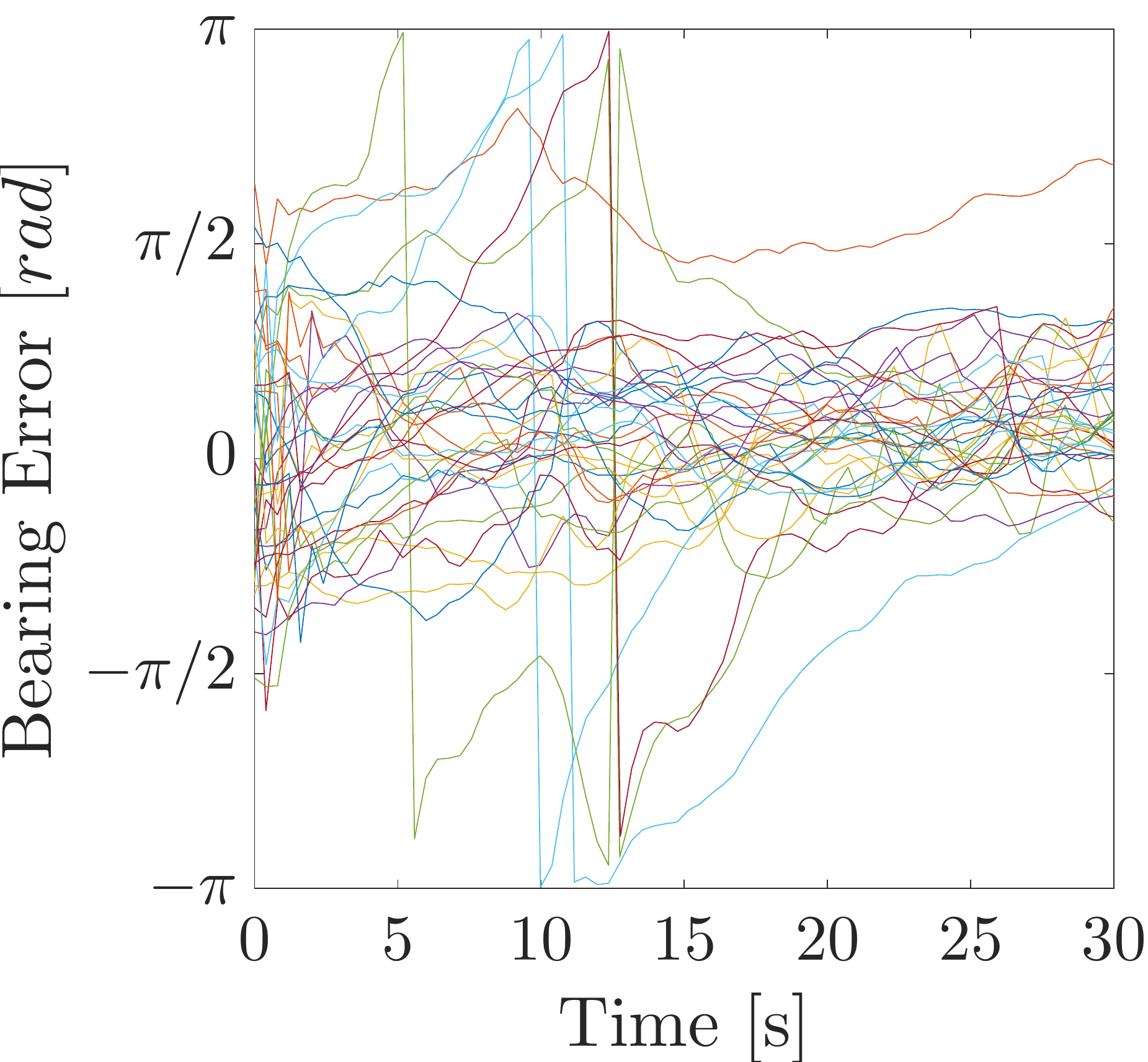}
      \caption{With three AR.Drones}
      \label{fig:error_convergence_3}
    \end{subfigure} 
    \caption{Comparison of bearing estimate errors in the first 30 seconds of flight during flights with external state measurements}
    \label{fig:error_convergence}
  \end{figure}

  All relative localization range errors shown in \figref{fig:rangeerror_controlled_2mav} and \figref{fig:rangeerror_controlled_3mav}, 
  and the bearing errors are shown \figref{fig:bearingerror_controlled_2mav} and \figref{fig:bearingerror_controlled_3mav}.
  The \gls{RMSE} for flights with two MAVs is $0.57rad$ for bearing, and $0.86m$ for range.
  With three MAVs, the \gls{RMSE} rises to $0.70rad$ and $1.14m$, respectively.
  On occasion, we observe that the bearing error temporarily diverges towards $\pm \pi$.
  This error does not necessarily lead to collisions due to the non-reciprocal nature of the avoidance behavior.
  Nevertheless, it introduces a temporary uncertainty in the system.
  The error is more frequent with three AR.Drones.
  We also observe that the convergence rate for bearing estimates over flights with three AR.Drones is worse than with two AR.Drones.
  This may be appreciated in \figref{fig:error_convergence}, zooming into the first $30s$ of \figref{fig:bearingerror_controlled_2mav} and \figref{fig:bearingerror_controlled_3mav} in more detail.
  Convergence times for flights with three MAVs reach up to $30s$ prior to settling (\figref{fig:error_convergence_3}).
  By comparison, the convergence in flights with two AR.Drones only (\figref{fig:error_convergence_2}) is found to be at most within $5-10s$.

\section{Experiments featuring On-board Own-State Measurements}
\label{sec:realworldexperiments_autonomous}
  The experiments from the controlled flights were repeated but with on-board state estimation by the MAVs.
  Therefore, on-board MAV sensors measured and controlled velocity, orientation, and altitude.
  This shows real-world relative localization performance for collision avoidance.
  \begin{figure*}[t!]
    \centering
    \begin{subfigure}[t]{84mm}
      \centering{}
      \includegraphics[width=\textwidth]{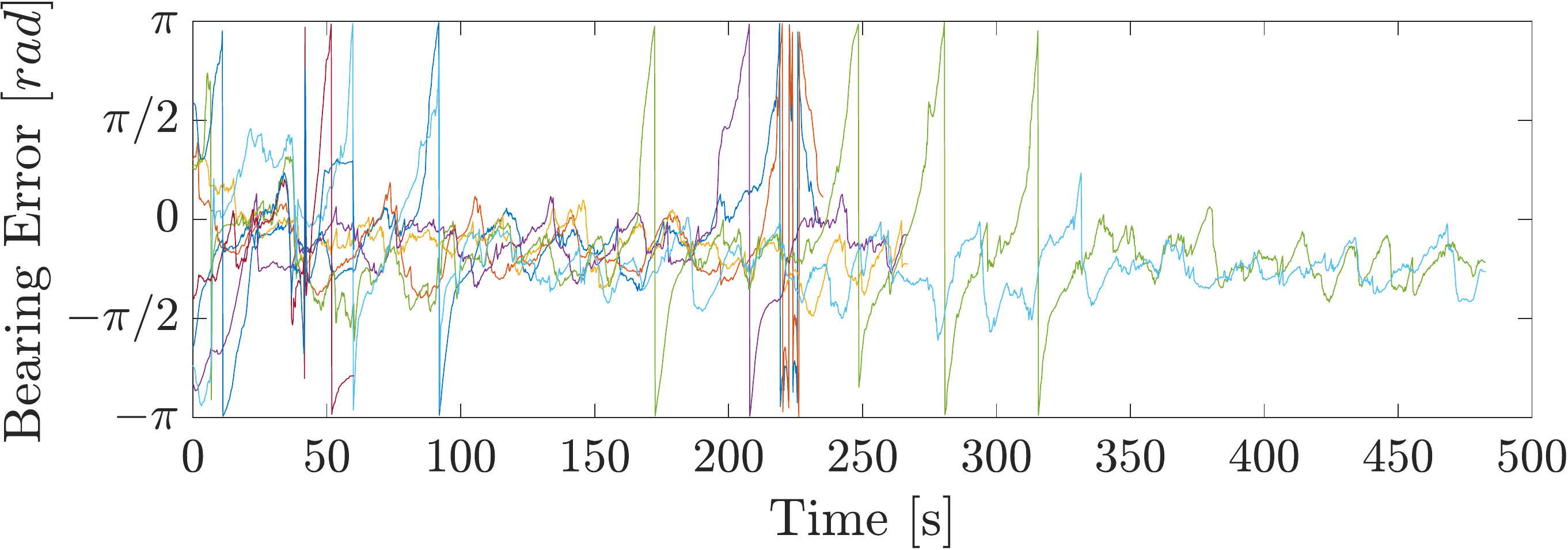}
      \caption{With two AR.Drones}
      \label{fig:bearingerror_autonomous_2mav}
    \end{subfigure} 
    ~
    \begin{subfigure}[t]{84mm}
      \centering
      \includegraphics[width=\textwidth]{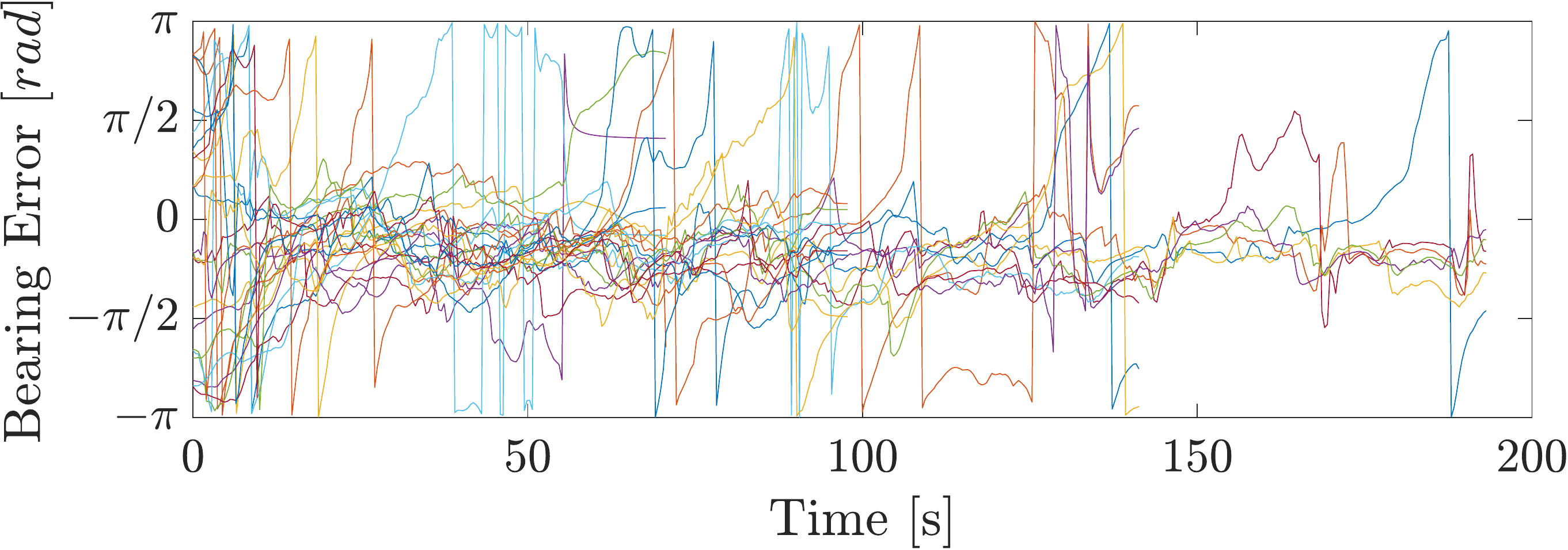}
      \caption{With three AR.Drones}
      \label{fig:bearingerror_autonomous_3mav}
    \end{subfigure}
    \caption{Overview of relative bearing estimation errors for flights with two AR.Drones (a) and three AR.Drones (b) featuring on-board state estimation}
    \label{fig:exp_errorplots_autonomous}
  \end{figure*} 
  
\subsection{Experimental Set-up}
\label{sec:realworldresults_autonomous_setup}
  Velocity was estimated using the bottom facing camera and the \emph{EdgeFlow} \citep{mcguire2016local}.
  Orientation was measured using gyroscope integration (given an initial orientation towards North).
  Height from the ground was measured using sonar.
  Optitrack was \emph{only} used to enforce condition M1 (wall detection), which featured $d_{safe}=0.5$
  This is because wall detection is outside of the purpose of this research.
  To further stress-test the system, a further change was that the EKFs initial relative position assumption was $x_{ji}=y_{ji}=1m$ for any MAV 
  $\mathcal{R}_i$ with respect to any other $\mathcal{R}_j$, as opposed to the center of the arena.
  The AR.Drones communicated with a ground-station using a Wi-Fi link for logging and take-off/land control.

\subsection{Results}
\label{sec:realworldresults_autonomous_results}
  Four flights were performed with two AR.Drones for a cumulative flight time of $17.3min$. 
  The flights lasted $3.9min$, $4.4min$, $8.0min$, and $1.0min$.
  Only the first and the last ended due to collisions.
  The second and third were ended due to low batteries.
  The third flight suffered from a near-collision in the early stages, but afterwards successfully continued until $8.0min$ without collisions.
  Another four flights were conducted with three AR.Drones, which lasted $8.3min$ cumulatively.
  The flights lasted $1.2min$, $3.2min$, $2.3min$, $1.6min$.
  The second flight was ended due to low batteries on one MAV.
  The other flights ended due to collisions between two of the three drones.\\

  The bearing estimation error is shown in 
  \figref{fig:exp_errorplots_autonomous}.
  The error has increased with comparison to the previous results.
  With two AR.Drones, the mean RMSE over the first three flights is $0.85rad$. This is sufficient for a long collision-free flight time.
  In the last flight, however, the RMSE was $1.3rad$, possibly due to a large disturbances in RSSI.
  This is eventually lead to a relatively early collision after $1.0min$.
  With three drones, the bearing RMSE over all flights is $1.0rad$.
  Furthermore, observing \figref{fig:exp_errorplots_autonomous} we can note an accumulating error bias in bearing over time due to the accumulating gyroscope bias.
  This should be corrected for future implementations by using the magnetometer to limit the accumulating bias.
  \begin{figure}[t!]
    \centering
      \includegraphics[width=84mm]{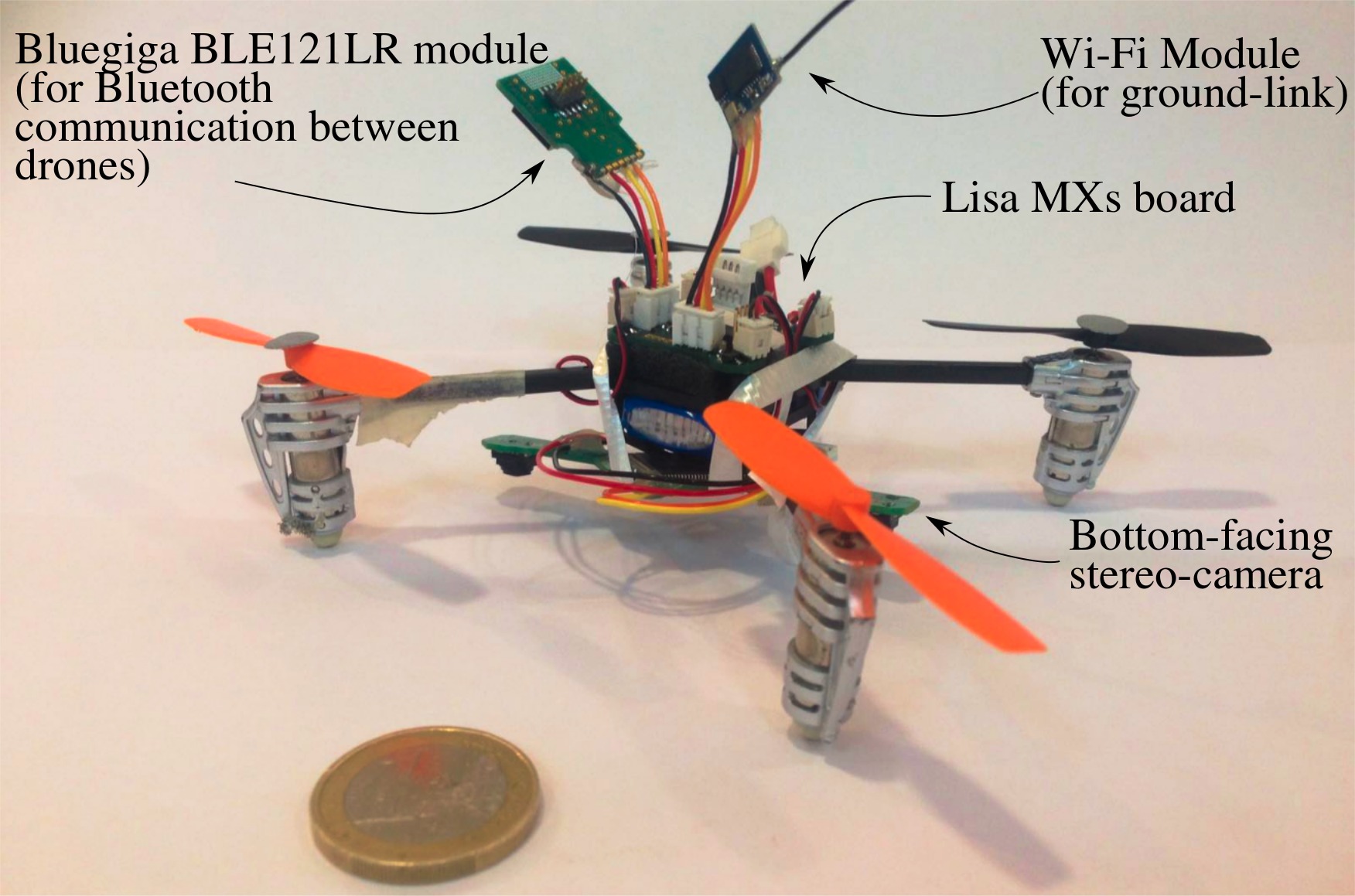}
      \caption{Miniature drone used in the experiments}
      \label{fig:pocketdrone}
  \end{figure} 

\section{Porting the technology to Miniature Drones}
\label{sec:miniaturedrones}
  To show that the proposed solution scales to smaller MAVs, we ported the technology to Ladybird MAV.
  One test-ready MAV and its components are shown in \figref{fig:pocketdrone}.
  As for the AR.Drones, the Wi-Fi link was used for logging and take-off/land control.

  \subsection{Experimental Set-Up} 
  \label{sec:miniaturedrones_setup}
	A bottom facing camera and a gyroscope measured velocity and orientation,  respectively.
	In the LD model, following a short hand-made calibration $P_n=-55dB$.
	Given the smaller size of the drones, the enforced flight arena was reduced to $2m\times2m$ with $d_{safe}=0.5m$.
	This scenario is similar to configuration 2 from \figref{fig:densitytable}.
  For simplicity, given the lack of a tested height sensor, height was controlled using Optitrack.
	The drones flew at slightly different heights to limit damages in case of failure by the collision avoidance system.
  All other test parameters stayed as for \secref{sec:realworldexperiments_autonomous}.

  \subsection{Results}
  \label{sec:miniaturedrones_results}
  	Three flights were performed with this set up, lasting $2.8min$, $3.7min$, and $3.1min$. 
    \footnote{The available flight footage shows how the MAVs avoided flying into each other for the duration of the flights. See \url{https://www.youtube.com/playlist?list=PL_KSX9GOn2P9f0qyWQNBMj7xpe1HARSpc}}
  	The first flight saw no collision cases.
  	The second flight saw near collisions at $1.35min$ and at $3.7min$.
    The latter came in light of low-batteries by one of the drones.
    As it lowered its height, the two MAVs also collided.
  	The third flight saw a near collisions after $\approx60s$ and $\approx90$. Both took place in the corner when condition M1 takes over the drones, and are thus are regarded more as a failure of the behavior than the relative localization.
  	This shows the importance of implementing a method that keeps taking into account other drones while also avoiding the walls, which was not implemented in our controller. \\

  	It is noted that a slightly lower performance than previous experiments was expected due to the smaller arena size, an effect which was also observed in simulation and is discussed further in 
  	\secref{sec:discussion_localizationperformance}.
  	Nevertheless, we also note a decrease in accuracy for relative localization as RMSE per flight ranges from $0.8rad$ to $1.37rad$.
  	Inspecting the data in more detail shows that this is the result of larger errors in both RSSI noise as well as lower quality on-board velocity estimates.
  	The former is explained by the fact that the Bluetooth module was placed right next to the Wi-Fi module, creating disturbances.

\section{Discussion}
\label{sec:discussion}

\subsection{Performance of Relative Localization}
\label{sec:discussion_fusionfilterperformance}
  In all AR.Drone tests, a noticeable loss in relative localization performance was measured when introducing a 3$^{rd}$ MAV.
  The effects were longer convergence times as well as higher relative bearing/range errors.
  A decrease in performance was also observed when using on-board velocity estimates.
  This was due to a combination of over-under estimation of velocity or occasional spikes in the measurements. \\

  The relative localization scheme was implemented with an \gls{EKF}.
  This may be criticized for its reliance on a Gaussian noise model.
  Robust \citep{kallapur2009discrete} or adaptive \citep{sasiadek1999sensor} variants of Kalman filters, or a \glspl{PF} \citep{svevcko2015distance}, might be better suited to this end.
  However, a mere change in filter could increase computational costs without bringing a higher quality estimate.
  This is because there are a number of other limitations. 
  \begin{itemize*}
  \item The logarithmic decrease in RSSI makes it intrinsically insufficient to measure changes in range at larger distances.
  Without measurable changes in RSSI, bearing is no longer observable.
  \item RSSI disturbances in the environment cannot be fully modeled unless the environment is known a-priori.
  \item The proposed process update equation makes the null assumption that all velocities remain constant between time-steps.
  Improvements may come from including more complex dynamic properties in the process equation, e.g. acceleration and/or jerk.
  \item As seen throughout our tests, major improvements can come by improving the quality of on-board state estimates.
  \end{itemize*}
  Further investigations are encouraged to define a filter that lowers the expected worst-case error.\\

  Further improvements could also come from a change in communication hardware.
  In this work, we have achieved promising results using Bluetooth, which was selected due to its prompt availability on several drones.
  The noise and disturbances with Bluetooth, however, are large.
  Other hardware, such as \gls{UWB}, would offer a significant reduction in noise, leading to better overall relative localization results.
  Based on our simulations from \secref{sec:discussion_varyingnoiselevels}, this should automatically result in an improved overall system performance.

\subsection{Performance of Collision Avoidance}
\label{sec:discussion_localizationperformance}
  \begin{figure*}[t!]
    \begin{adjustbox}{minipage=174mm,scale=1.0}
    \begin{subfigure}[t]{0.333\columnwidth}
      \centering
      \includegraphics[width=\textwidth]{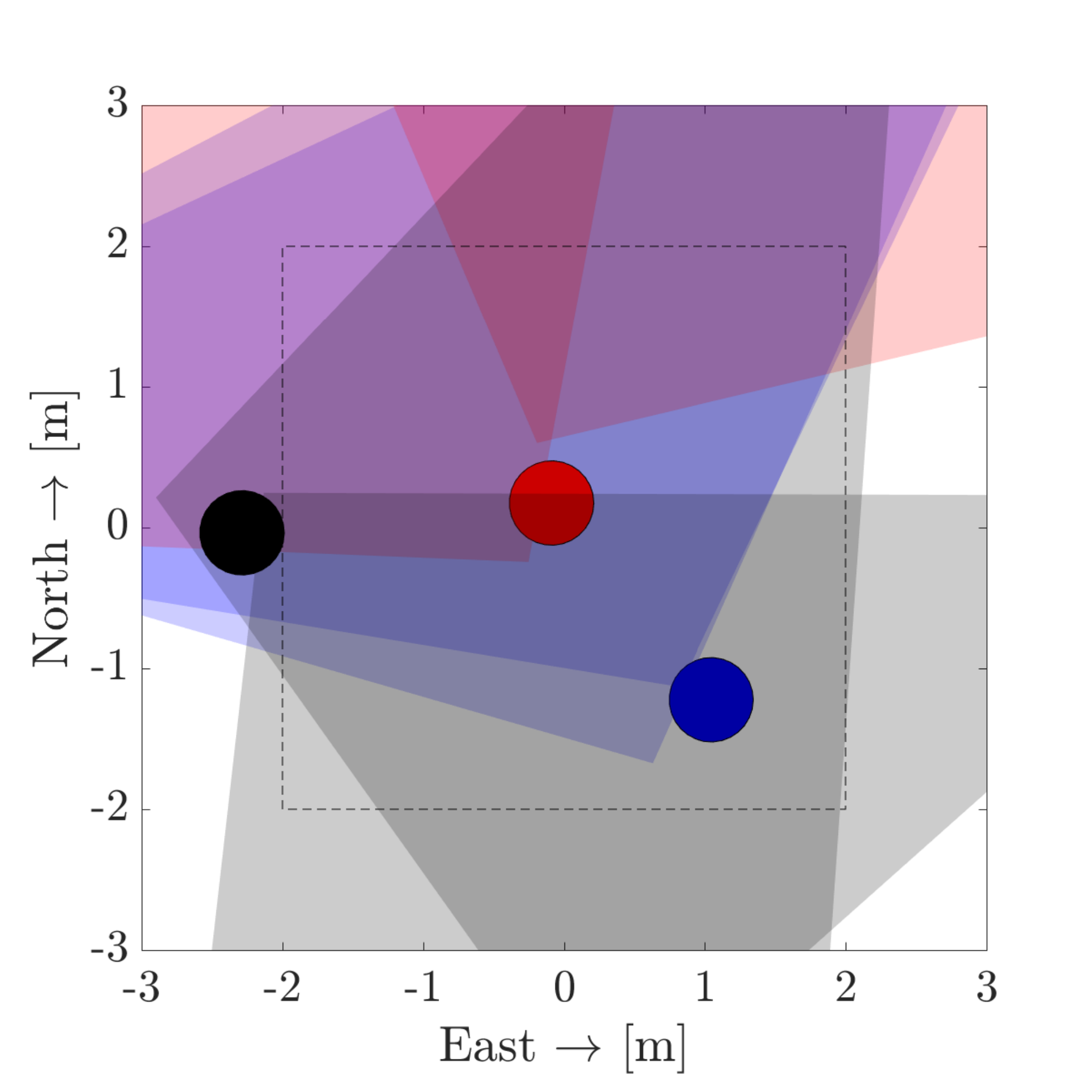}
      \caption{Time = 180s}
      \label{fig:collisioncase_screenshot_1}
    \end{subfigure} 
    ~
    \begin{subfigure}[t]{0.333\columnwidth}
      \centering
      \includegraphics[width=\textwidth]{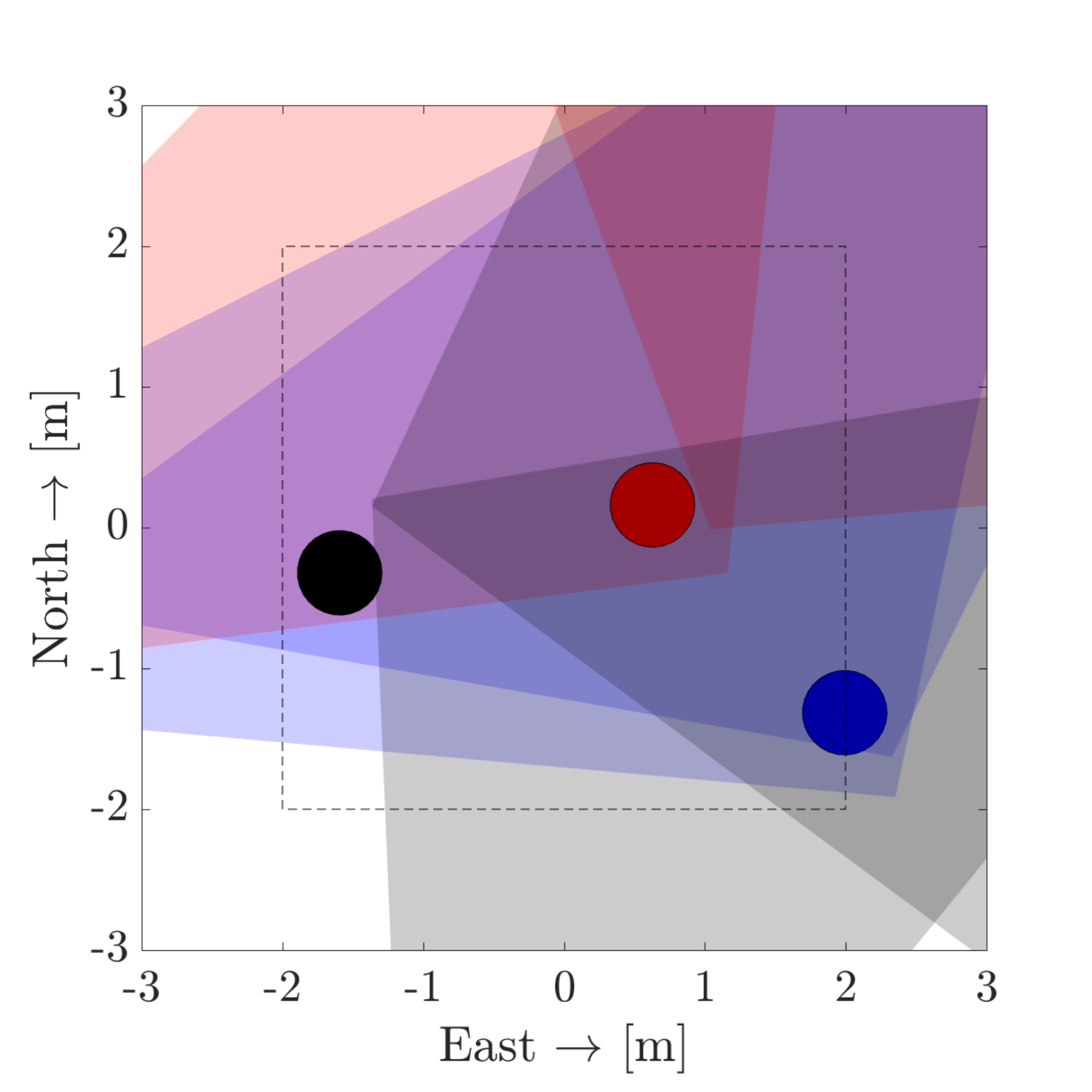}
      \caption{Time = 182s}
      \label{fig:collisioncase_screenshot_2}
    \end{subfigure}
    ~
    \begin{subfigure}[t]{0.333\columnwidth}
      \centering
      \includegraphics[width=\textwidth]{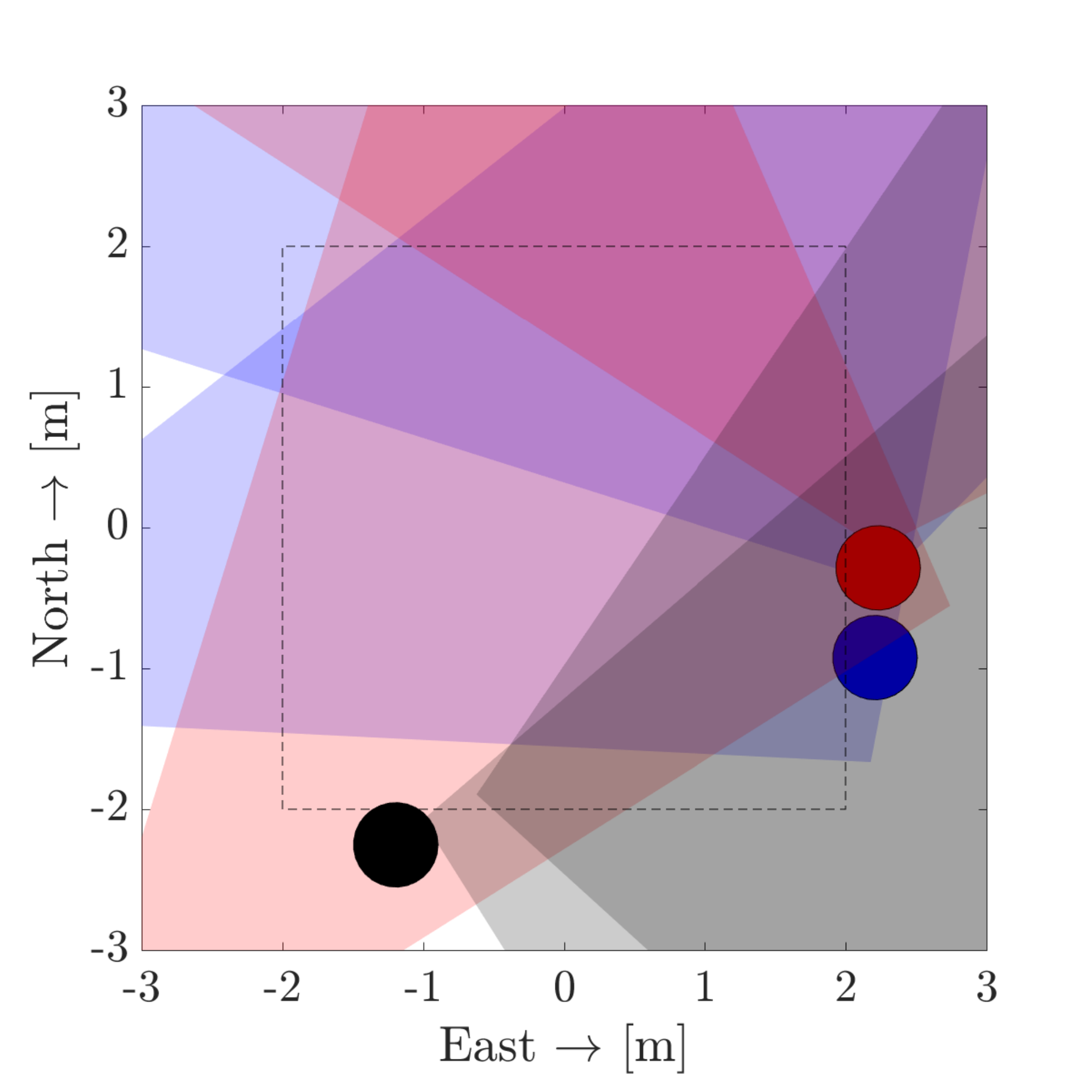}
      \caption{Time = 188s}
      \label{fig:collisioncase_screenshot_3}
    \end{subfigure}
    \end{adjustbox}
    \caption{
    Chronological depiction (left to right) of a collision case in a flight with 3 AR.Drones.
    Large circles indicate the ground-truth position in the arena, the triangles are the cones that each AR.Drone holds.
    }
    \label{fig:collisioncase_screenshot}
  \end{figure*}
  In simulation, all twelve configurations have also been tested \emph{without} active collision avoidance.
  The obtained mean flight times ranged between $3.9s$ and $14.3s$.
  A $z$-test with $95\%$ confidence level \citep{dekking2005modern} shows a statistically significant improvement in flight time for all configurations when using our method.\\

  \figref{fig:flighttimedensity} showed that smaller rooms lead to poorer performance than larger rooms despite similar airspace density.
  The parameter $\varepsilon_\alpha$, as explained in \secref{sec:preservingbehavior}, implements room scaling within the collision cones.
  Reasons for this are:
  \begin{itemize*}
    \item The ratio of arena size to $v_{nominal}$ decreases in smaller rooms.
    \item The communication rate is constant, which limits the decision rate of the collision avoidance controller.
    \item In smaller rooms, M1 is called more frequently, in which case collision cones are ignored according to the task in this article.
  \end{itemize*}

  It was observed, and confirmed in simulation, that collisions for flights with three MAVs likely occur along the edges of the area.
  In the simulations of configuration 11, which is the one tested with the AR.Drones, $81\%$ of the collided simulated flights with three MAVs ended within $0.5m$ of the arena borders.
  By comparison, only $35\%$ of collisions with two MAVs occurred within this space. 
  An example extracted from an AR.Drone flight recounted by the three events below (shown in \figref{fig:collisioncase_screenshot}).
  \begin{enumerate*}
    \item One MAV is at the corner and reluctant to make movements towards the center.
    At time $t=180s$ (\figref{fig:collisioncase_screenshot_1}), we see this for the bottom right AR.Drone (blue).
    Its slow speed causes the red MAV to mistaken its estimate of the blue drone.
    In normal conditions, collision avoidance could still be achieved by the blue MAV, but it cannot react as it is trapped in the corner. 
    \item Another MAV turns towards the same side.
    In time $t=182s$ (\figref{fig:collisioncase_screenshot_2}), the central AR.Drone (red) avoids the black AR.Drone (on left) but in doing so goes to the right.
    \item The second MAV also ends along the border and reluctant to make movements.
    At time $t=188s$ (\figref{fig:collisioncase_screenshot_3}), the two oscillate along the border until a collision occurs.
  \end{enumerate*}
  This scenario is less likely with two MAVs due to the larger freedom of movement and the higher relative localization accuracy.
  One method to limit this would be to reduce the angle of the collision cones for further-away MAVs, which increases mobility.
  Furthermore, it is also necessary to create an avoidance scheme that takes into account the wall and the drones together. 
  This shall be tackled in future work.

\section{Conclusion and Future Work}
\label{sec:conclusion}
We have shown that it is possible to use wireless communication as a relative localization sensor that can be used on-board of MAVs operating in a team.
This leads to a large reduction in collisions \emph{without the need of a dedicated sensors}.
With the solution proposed in this paper, a team of AR.Drones in a $4m\times4m$ area could fly for several minutes without collisions.
The technology was also used with miniature drones, showing its portability.
With respect to the scenario in mind (i.e. the exploration of indoor spaces by MAV teams), this is an efficient method to limit collision risks in the event that MAVs end up flying in the same room.

The combined relative localization/collision avoidance system as presented and tested in this paper will be further improved in future work.
Importantly, we will investigate \gls{UWB} modules instead of a Bluetooth modules.
Bluetooth suffers from high disturbances and noise that is detrimental to the performance, especially as more MAVs are introduced.
Using \gls{UWB} instead is expected to considerably improve the distance measurements used by the filter.
Furthermore, the introduction of an avoidance strategy that makes a more informed decision near walls or when multiple MAVs are present is needed.
This could resolve the more complex collision scenarios, especially in smaller rooms.

\appendix
\section*{Videos}
\label{sec:appendices}
Videos of experiments are available at: \url{https://www.youtube.com/playlist?list=PL_KSX9GOn2P9f0qyWQNBMj7xpe1HARSpc}.

\bibliographystyle{spbasic}   
\bibliography{main.bib}   

\end{document}